\documentclass[11pt]{article}

\usepackage[final]{acl}

\usepackage{times}
\usepackage{latexsym}

\usepackage[T1]{fontenc}

\usepackage[utf8]{inputenc}

\usepackage{microtype}

\usepackage{inconsolata}

\usepackage{graphicx}

\usepackage{amsmath,amsfonts,bm}
\usepackage{bbm}

\def\eqref#1{equation~\ref{#1}}

\def\1{\bm{1}}

\DeclareMathAlphabet{\mathsfit}{\encodingdefault}{\sfdefault}{m}{sl}
\SetMathAlphabet{\mathsfit}{bold}{\encodingdefault}{\sfdefault}{bx}{n}

\usepackage{multirow}
\usepackage{booktabs}
\usepackage{graphicx}
\usepackage{wrapfig}
\usepackage{placeins}
\usepackage{array}
\usepackage{longtable}

\usepackage{xcolor}
\usepackage[most]{tcolorbox}

\newtcolorbox{codeblock}{
  enhanced,
  colback=black!2,
  colframe=black!20,
  boxrule=0.4pt,
  arc=2pt,
  left=2mm,right=2mm,top=1mm,bottom=1mm,
  fontupper=\ttfamily\small,
  breakable
}

\newcommand{\tok}[1]{%
  \texttt{\textcolor{black!55}{<}\textcolor{green!50!black}{#1}\textcolor{black!55}{>}}
}
\newcommand{\opentag}[1]{%
  \texttt{\textcolor{black!55}{<}\textcolor{blue!70!black}{#1}\textcolor{black!55}{>}}%
}
\newcommand{\closetag}[1]{%
  \texttt{\textcolor{black!55}{</}\textcolor{blue!70!black}{#1}\textcolor{black!55}{>}}%
}

\newtcolorbox{windowbox}[2][]{%
  enhanced,
  breakable,
  colback=white,
  colframe=black,
  boxrule=0.8pt,
  arc=2mm,
  left=3mm,right=3mm,top=2mm,bottom=2mm,
  fonttitle=\bfseries,
  coltitle=white,
  colbacktitle=black,
  title={#2},
  attach boxed title to top left={xshift=0mm,yshift=-1mm},
  boxed title style={sharp corners, boxrule=0pt, interior style={fill=black}},
  #1
}

\newtcolorbox{fieldbox}[2][]{%
  enhanced,
  breakable,
  colback=gray!3,
  colframe=black!25,
  boxrule=0.5pt,
  arc=1.5mm,
  left=2mm,right=2mm,top=1mm,bottom=1mm,
  fonttitle=\bfseries,
  title={#2},
  listing only,
  listing options={
    basicstyle=\ttfamily\footnotesize,
    columns=fullflexible,
    breaklines=true,
    showstringspaces=false,
  },
  #1
}

\title{MMAI Gym for Science: \\ Training Liquid Foundation Models for Drug Discovery}

\author{Maksim Kuznetsov\textsuperscript{1}, Zulfat Miftahutdinov\textsuperscript{1}, Rim Shayakhmetov\textsuperscript{1}, Mikolaj Mizera\textsuperscript{1}, Roman Schutski\textsuperscript{1} 
\\
{\bf Bogdan Zagribelnyy\textsuperscript{1}}, {\bf Ivan Ilin\textsuperscript{1}}, {\bf Nikita Bondarev\textsuperscript{1}}, {\bf Thomas MacDougall\textsuperscript{1}}, {\bf Mathieu Reymond\textsuperscript{1}}
\\ 
{\bf Mihir Bafna\textsuperscript{2}}, {\bf Kaeli Kaymak-Loveless\textsuperscript{2}}, {\bf Eugene Babin\textsuperscript{1}}, {\bf Maxim Malkov\textsuperscript{1}}, {\bf Mathias Lechner\textsuperscript{2}}
\\
{\bf Ramin Hasani\textsuperscript{2}}, {\bf Alexander Amini\textsuperscript{2}}, {\bf Vladimir Aladinskiy\textsuperscript{1}}, {\bf Alex Aliper\textsuperscript{1}}, {\bf Alex Zhavoronkov\textsuperscript{1}}
\AND 
\textsuperscript{1} \raisebox{-0.25\height}{%
  \includegraphics[height=1.3em]{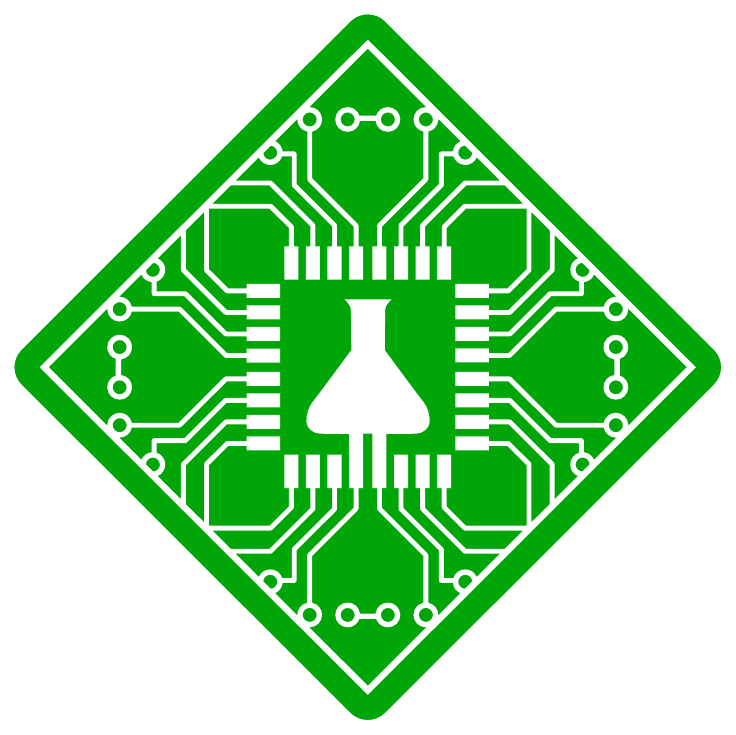}%
} Insilico Medicine \And \textsuperscript{2}  \raisebox{-0.25\height}{%
  \includegraphics[height=1.3em]{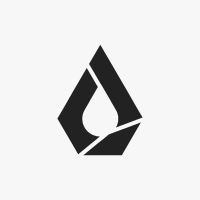}%
} Liquid AI
}

\begin{document}

\maketitle

\begin{abstract}
General-purpose large language models (LLMs) that rely on in-context learning do not reliably deliver the scientific understanding and performance required for drug discovery tasks. Simply increasing model size or introducing reasoning tokens does not yield significant performance gains. To address this gap, we introduce the MMAI Gym for Science, a one-stop shop molecular data formats and modalities as well as task-specific reasoning, training, and benchmarking recipes designed to teach foundation models the ''language of molecules'' in order to solve practical drug discovery problems. We use MMAI Gym to train an efficient Liquid Foundation Model (LFM) for these applications, demonstrating that smaller, purpose-trained foundation models can outperform substantially larger general-purpose or specialist models on molecular benchmarks. Across essential drug discovery tasks - including molecular optimization, ADMET property prediction, retrosynthesis, drug–target activity prediction, and functional group reasoning - the resulting model achieves near specialist-level performance and, in the majority of settings, surpasses larger models, while remaining more efficient and broadly applicable in the domain.

\end{abstract}

\section{Introduction}
\label{sec:intro}

Large language models (LLMs) are increasingly being explored as tools for drug discovery~\citep{taylor2022galactica, pei2023biot5, narayanan2025ether0}, either by fine-tuning general-purpose open-source models~\cite{wang2025txgemma} or by training domain-specific models for individual downstream tasks, such as property prediction, synthesis planning, and target prioritization~\citep{edwards2022molt5, livne2024nacho}. Despite promising demonstrations, current LLM-based approaches typically lag behind specialist methods on core drug discovery tasks (e.g, e.g., ADME and toxicity prediction, retrosynthesis~\citep{zagribelnyy2026chemcensor}), and performance gains obtained via prompting or task-local fine-tuning often do not translate into robust, cross-task capability. Prompt engineering or naive fine-tuning typically yields limited robustness, especially under out-of-distribution shifts. As a result, it is challenging to obtain a single "generalist" model that performs competitively in the diverse molecular reasoning workloads encountered in medicinal chemistry, biology, and early clinical development.

In this work, we propose a training recipe that uses  supervised  (SFT) and reinforcement learning (RFT) fine-tuning on domain-specific data to turn a general-purpose causal LLM into a drug-discovery generalist. We instantiate this approach in \textbf{MMAI Gym}, a structured training and evaluation environment that provides curated scientific reasoning traces across key modalities and tasks in drug discovery. Unlike approaches that rely primarily on generic reasoning reinforcement learning (RL) or fine-tuning on a single benchmark’s training split, MMAI Gym emphasizes domain-faithful reasoning chains, task formats used by practitioners, and evaluation under distribution shift via held-out and out-of-distribution benchmarks. 

We demonstrate our approach by adapting the Liquid Foundation Model~\cite{amini2025lfm2}, an efficient language model architecture, using MMAI Gym. Our results show that a single SFT+RFT run substantially improves - in some cases by up to an order of magnitude - LFM2's performance on drug discovery-oriented benchmarks. The resulting models achieve competitive or state-of-the-art performance relative to specialist and few-shot-prompted foundation models across diverse drug discovery tasks, including molecular property prediction,  optimization, functional-group reasoning, retrosynthesis, and unconditional and pocket-conditioned 3D molecular generation. These results suggest that domain-specific reasoning supervision can be an effective for scientific LLMs.

\begin{figure*}[t]
  \centering
  \vspace{-1em}
  \includegraphics[width=\linewidth]{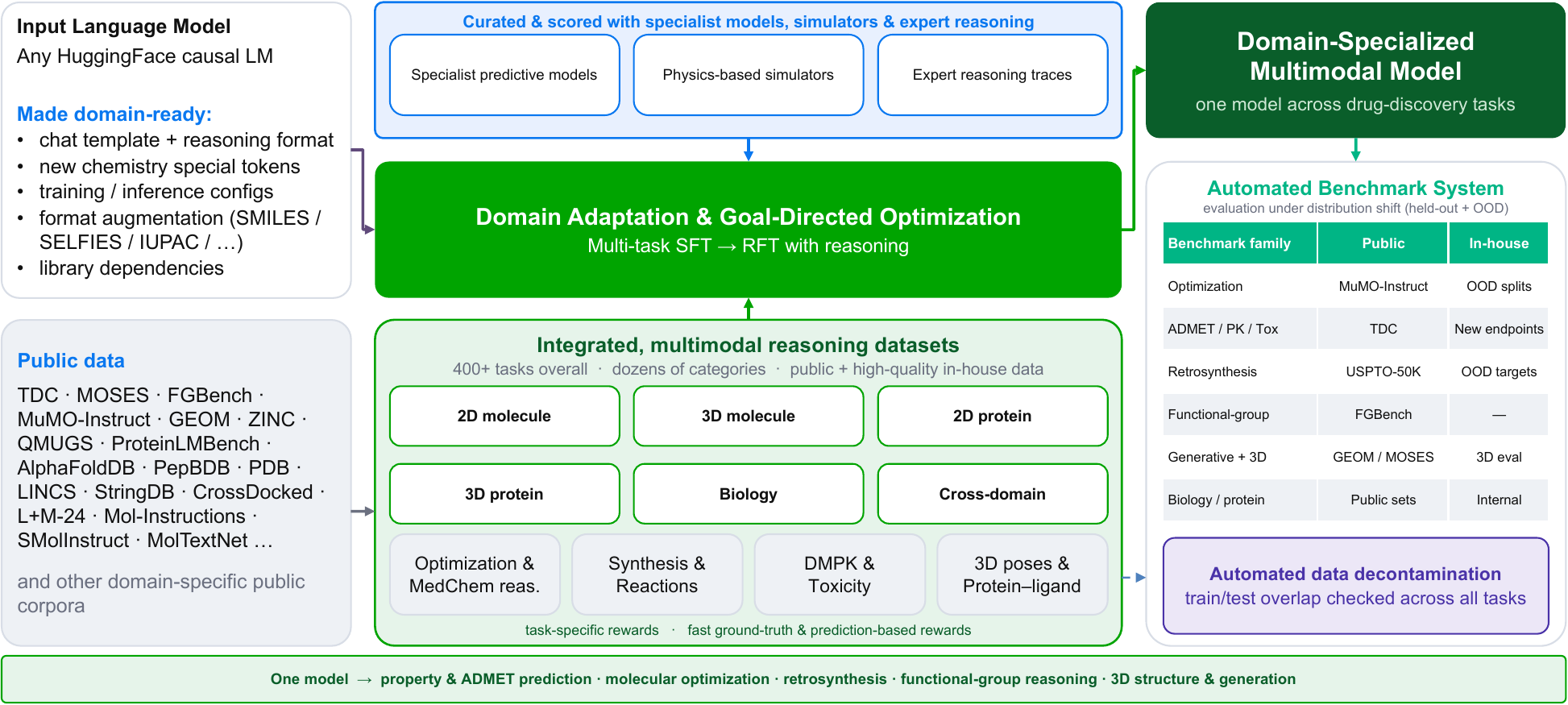}
  \caption{\textbf{MMAI Gym}: Integrated Data, Training, and Benchmark Suite}
  \label{fig:mmai_gym}
  \vspace{-1em}
\end{figure*}

\section{MMAI Gym for Science}
\label{sec:mmai_gym}

Deploying LLMs in chemistry, biology, and other domains requires the ability to handle very long context inputs efficiently, to perform robustly across a diversity of highly specialized downstream tasks (i.e., from predicting scalar toxicity values to designing experimental protocols), and to generalize to unseen scenarios and distribution shifts (i.e., new chemo-types and novel protein targets).
We thus hypothesized that a domain-specific training approach would enable the efficient, long-context LFM2 architecture (see \autoref{app:lfm}) to achieve strong performance in scientific scenarios.

To demonstrate this in the context of chemical drug discovery, we built and leveraged MMAI (Multi-Modal AI) Gym for Science\footnote{\href{https://insilico.com/mmai}{MMAI Gym for Science website}}, a suite of data, training, and evaluation environment designed to adapt and improve general-purpose and chemical domain-specific LLMs to molecular reasoning tasks primarily in drug discovery and development (Figure \ref{fig:mmai_gym}). MMAI Gym treating drug discovery problems as domain-specific reasoning tasks rather than pure text prediction, targeting the formats and causal chains used by medicinal chemists, biologists, and clinical researchers.

MMAI Gym implements a multi-stage curriculum combining (i) curated, domain-specific reasoning datasets spanning medicinal chemistry optimization chains, synthesis and reaction reasoning, drug metabolism, pharmacokinetics (DMPK), toxicity modeling, and geometry-grounded tasks incorporating 3D structures; (ii) multi-task supervised and reinforcement fine-tuning; and (iii) automated data decontamination plus evaluation on public and internal out-of-distribution benchmarks. The intended outcome is a single adapted model that can jointly perform property prediction, hypothesis  design/optimization, synthesis planning, target/biology reasoning, and select clinical inference tasks with improved performance and reliability.  

\subsection{Datasets and Data Balancing}

MMAI Gym comprises a suite of $400+$ tasks spanning dozens of categories, the main high-level ones being 2D molecule tasks, 3D molecule tasks, 2D protein tasks, 3D protein tasks, drug–gene interaction tasks, and cross-domain tasks that operate jointly in multiple modalities. Each broad category covers a variety of task formulations, including predictive tasks with easily verifiable answers and molecular generation tasks where a large set of diverse answers is possible.

The molecular tasks are built from a range of datasets, including but not limited to TDC~\citep{huang2021tdc}, MOSES~\citep{polykovskiy2020moses}, FGBench~\citep{liu2025fgbench}, MuMO-Instruct~\citep{dey2025mumoinstruct}, CREED~\citep{zagribelnyy2026chemcensor}, and MolTextNet~\citep{zhu2025moltextnet}-and the 3D molecular (spatial) tasks draw on GEOM~\citep{axelrod2022geom}, ZINC~\citep{irwin2005zinc}, and QMUGS~\citep{isert2022qmugs}. Protein tasks leverage ProteinLMBench~\citep{shen2024proteinlmbench}, the AlphaFold Protein Structure Database~\citep{varadi2023alphafolddb}, CREMP~\cite{grambow2024cremp} and the BC40 subset of PDB~\citep{berman2000pdb}. Interaction tasks are based on StringDB~\cite{szklarczyk2025stringdb}, while cross-domain tasks incorporate CrossDocked2020~\cite{francoeur2020crossdocked}, L+M-24~\cite{edwards2024lm24}; across main categories, these datasets are complemented with high-quality in-house data. Additional details along with datasets statistics can be found in \autoref{app:datasets}.

To integrate the diverse datasets and modalities, we adopt a multi-stage balancing strategy. When constructing training batches, we first sample uniformly from the six top-level categories; within the selected category, we then sample a dataset-task uniformly and finally draw an example from that dataset. This procedure is designed to mitigate imbalances arising from differing dataset sizes across modalities and sources, ensuring that all tasks are represented uniformly within each category.

\subsection{Tokenization and Augmentation of Chemical Entities}
To enable LFM2 model to learn abstract molecular concepts as opposed to overfitting to a particular syntax of the molecular format, we use chemical format conversion as part of our data augmentation strategy. For each user input, we randomly convert SMILES~\cite{weininger1988smiles} to SELFIES~\cite{krenn2020selfies} and vice versa; additionally, SMILES/SELFIES strings are randomly converted to IUPAC names when possible. This format augmentation helps the model to learn modality-independent molecular concepts and acts as linguistic bridge between valid chemistry and natural language. To further enhance chemical generalization, we also augment SMILES and SELFIES representations during training by applying non-canonical random traversal both for user input and model output.

Following prior chemical language models~\cite{livne2024nacho, pei2023biot5}, we augment LFM2's base natural language vocabulary with tokens specific to SMILES, SELFIES, and FASTA formats. This separation is intended to distinguish chemical symbols and ensure a consistent encoding of molecular strings. We additionally wrap each molecular sequence with format type tags. For example, an Acetic Acid is isolated as \verb|<smiles>CC(=O)O</smiles>| and tokenized as :

\begin{codeblock}
[\opentag{smiles},\tok{sm\_C},\tok{sm\_C},\tok{sm\_(},\tok{sm\_=},\tok{sm\_O},\\

\tok{sm\_)},\tok{sm\_O},\closetag{smiles}]
\end{codeblock}

During training, we enforce this specialized format isolation and tokenization for model-generated outputs, while applying it to user inputs with probability $0.5$ to better align the base model representations with the newly introduced molecular tokens. This allows the model to understand and convert user input that contains only standard text tokens, while having the power to represent the molecules for best performance.

To represent 3D molecular and protein structures, we follow the text-based representations introduced in nach0-pc~\cite{kuznetsov2024nachopc} and BindGPT~\cite{zholus2024bindgpt}. For small molecules, we first provide the molecular graph as a SMILES string, and then list the 3D coordinates of each atom in the same order as in the SMILES. For proteins, we lay out atomic coordinates residue-by-residue, annotating each residue with a dedicated amino-acid token (e.g., \texttt{\tok{am\_A}} for alanine, \texttt{\tok{am\_R}} for arginine, \ldots, \texttt{\tok{am\_V}} for valine) and each atom with an atom-name token (e.g., \texttt{\tok{atom\_name\_CA}}, \texttt{\tok{atom\_name\_CB}}, \ldots, \texttt{\tok{atom\_name\_NH2}}). To optimize textual context usage, we omit hydrogens and reconstruct their positions post hoc using standard cheminformatics toolkits~\cite{oboyle2011openbabel, landrum2023rdkit}.

\subsection{Reinforcement Learning Rewards}

All MMAI Gym tasks can be classified as 3 different kinds of tasks: (i) regression, (ii) classification, (iii) generation. To perform reinforcement learning post-training, each type incorporates a reward function, providing a score for the generated completion. For regression, we compute the mean absolute error between the prediction and the ground truth. Classification uses a binary reward signal, while generation computes molecular validity. Additionally, MMAI Gym includes 2 generic reward functions applicable to all tasks: a format reward to wrap thinking in \tok{think} tags, and a thinking reward to encourage longer thinking/reasoning sequences. More details about all reward functions are profided in \autoref{app:rl_rewards}.

During our experiments, We find it is important that all reward functions share the same scale, so that they do not interfere with each other during multi-task training. Moreover, due to the wide variety of tasks, we keep a high KL-regularization weight of 0.4 to ensure stable training.

Online RFT is performed using Group Relative Policy Optimization (GRPO)~\cite{shao2024deepseekmathpushinglimitsmathematical}, using the same tasks and prompts as in SFT. More details about the hyperparameters used can be found in \autoref{app:training}.

\section{Related Work}
\label{sec:related_work}

\paragraph{Molecular Graphs Reasoning and Generation.}
Tasks defined over molecular graphs or string representations span property prediction, structural reasoning, molecular optimization, and synthesis planning. Functional-group reasoning examines how local structural changes affect molecular properties, while multi-objective optimization seeks structural modifications that improve several potentially conflicting properties without departing excessively from the starting molecule. ADME, pharmacokinetic, and toxicity prediction map molecular structure to experimentally relevant endpoints, whereas retrosynthesis recovers plausible reactants for a target product. We evaluate these capabilities using FGBench~\cite{liu2025fgbench} and MuMO-Instruct~\citep{dey2025mumoinstruct}, TDC~\citep{huang2021tdc}, and  ChemCensor~\citep{zagribelnyy2026chemcensor}.

\paragraph{Spatial Molecular Generation.}
Unlike graph-based tasks, spatial molecular tasks require models to reason not only about molecular connectivity, but also about three-dimensional poses, molecular geometry, and the spatial arrangement of atoms and functional fragments. Unconditional 3D generation aims to produce diverse and chemically valid molecules that reproduce realistic distributions of bond lengths, angles, dihedrals, and conformations \cite{peng2023moldiff}. Pocket-conditioned generation additionally requires designing ligand structures and poses compatible with a given protein binding site~\cite{peng2022pockettomol, macdougall2026threedfit}.

\paragraph{Specialist Chemical Models and General-Purpose Foundation Models.}
Chemical downstream tasks have traditionally been addressed by specialist models trained for a narrow capability. These include chemistry-focused LLMs such as ChemLLM~\cite{Zhang2024ChemLLM}, NatureLM~\cite{Xia2025NatureLM}, nach0~\citep{livne2024nacho}, and TxGemma~\citep{wang2025txgemma}, as well as task-specific non-LLM methods. Recently, general-purpose foundation models from the GPT~\cite{openai2025gpt52}, Claude~\cite{anthropic2025claudeopus45}, Gemini~\cite{google2025gemini3flash}, and Grok~\cite{xai2025grok41} families have demonstrated increasingly strong chemical reasoning and generation capabilities, becoming competitive with specialist systems on several downstream benchmarks. Our work follows this generalist direction by adapting a single foundation model to tasks spanning molecular graphs, chemical reactions, and three-dimensional molecular structures.

\section{Experiments}
\label{sec:experiments}

We trained the domain-adapted models\footnote{\href{https://www.liquid.ai/request-access-mmai}{LFM2-MMAI} models}, LFM2-2.6B-MMAI and LFM2-24B-A2B-MMAI, on the MMAI Gym datasets and evaluated it on a variety of cheminformatic and drug discovery tasks, comparing against several baseline tiers. Training is performed in two stages: first using supervised fine-tuning (SFT), then applying reinforcement learning fine-tuning (RFT) in both (i) multi-domain multi-task (\textbf{MT}) and (ii) single-category or, simply, single-task (\textbf{ST}) settings. Throughout the experiments, \textbf{ST} denotes single-category training, even when a category contains multiple tasks; for example, the ADMET category comprises 22 property-prediction tasks. Full training details are provided in \autoref{app:training}.

\subsection{Benchmarking Protocol}
\label{sec:benchmarking}

For each benchmark task, every test example is evaluated multiple times. At each repetition, a prompt is assembled by randomly sampling from a pool of paraphrased instruction templates and the molecular input is independently augmented according to our augmentation and tokenization protocol, yielding a distribution of predictions over diverse prompt formulations, input augmentations, and stochastic decoding samples. The model generates a completion with a chain-of-thought reasoning trace enclosed in \tok{think} tags, followed by an \tok{answer} block. A parser extracts the prediction from the \tok{answer} block, either a number for regression tasks, a class label for classification tasks, or a molecule for generation tasks. When available, we additionally extract class probabilities from the first-token logprobs via a softmax over the tokens representing each class for metrics requiring class probabilities. The resulting distribution of predictions across repetitions is aggregated per example. When available, we use a task-specific aggregation method if the benchmark requires it. In other cases, by default, we use the median for continuous values and the majority vote for categorical ones.

Comparisons with larger proprietary models are asymmetric: LFM2-MMAI is trained on the relevant task distributions, whereas proprietary baselines use few-shot prompting without training. For base models not fine-tuned on our instruction format, we use author-provided prompts or explicit instructions with few-shot examples randomly sampled from the corresponding training split.

\begin{table}[t]
\centering
\resizebox{0.89\linewidth}{!}{
    \begin{tabular}{l|ccccc}
\toprule
\textbf{Model} & \textbf{BDP} & \textbf{BDQ} & \textbf{BPQ} & \textbf{DPQ} & \textbf{BDPQ} \\
\midrule
\multicolumn{6}{c}{\textit{Proprietary LLMs}} \\
\midrule
GPT-5.5 & 83.2 & \textbf{91.6} & 92.6 & 72.4 & \textbf{77.0} \\
Claude 4.6 Opus & 69.4 & 62.2 & 98.6 & \textbf{73.2} & 56.4 \\
\midrule
\multicolumn{6}{c}{\textit{Task-specific LLMs}} \\
\midrule
$\mathrm{GeLLM}^{3}\,\mathrm{O}\text{-}3_{\mathrm{Mistral}}$ & 84.8 & 87.0 & 93.0 & 62.8 & -- \\
$\mathrm{GeLLM}^{3}\,\mathrm{O}\text{-}3_{\mathrm{Llama}}$ & \textbf{86.8} & 90.0 & 94.0 & 60.6 & -- \\
\midrule
\multicolumn{6}{c}{\textit{Liquid Foundation Models}} \\
\midrule
2.6B-MMAI, MT & 86.0 & \textbf{91.6} & \textbf{98.8} & 71.2 & 56.4 \\
2.6B-MMAI, ST & 86.0 & 91.2 & \textbf{98.8} & 69.8 & 57.8 \\
24B-A2B-MMAI, ST & 68.2 & 78.8 & 97.2 & 60.2 & 60.4 \\
\bottomrule
\end{tabular}
}
\caption{\textbf{Success Rate on Molecular Optimization results on MuMO-Instruct.}}
\label{tab:mumo_res}
\end{table}

\subsection{Experimental Results and Discussion}
\label{sec:exp_discussion}

For consistency and relevance to drug discovery applications, we focus our experiments on widely adopted benchmark tasks spanning both 2D and 3D molecular reasoning: multi-objective optimization, functional-group reasoning, single-step retrosynthesis, ADMET/PK and toxicity prediction, spatial molecular distribution learning, and pocket-conditioned 3D ligand generation. Due to space constraints, we present results for a representative subset of baselines and key metrics in the main text. Comprehensive comparisons across all baselines and metrics\footnote{Newly released baselines on \href{https://dddbench.insilico.com}{DDDBench}}, along with details such as the number of runs and samples, are provided in \autoref{app:ext_experiments}.

\paragraph{Multi-objective molecular optimization.} Table~\ref{tab:mumo_res} shows that MMAI Gym substantially strengthens multi-objective molecular optimization on MuMO-Instruct. The adapted LFM2 models achieve the strongest and most consistent Success Rate performance across the three-property tasks, and remain among the top performers on the more difficult four-property setting. It is important to note that while some baselines preserve a higher similarity to the input molecule, this often comes at the cost of lower optimization success, and vice versa. For example, some baselines produce extreme relative improvement scores but with similarity values of lower than $0.2$, completely violating an instruction to keep the output similar to the input. In contrast, the MMAI-trained models strike a better balance between making effective structural edits and satisfying the target property constraints. While a smaller dense LFM2-2.6B-MMAI model outperforms LFM2-24B-A2B-MMAI on average, the MoE architecture handles the complex case with most conditions better.

\begin{table}[t]
\centering
\resizebox{1\linewidth}{!}{
    \begin{tabular}{l|cc|cc|cc}
\toprule
\multirow{2}{*}{\textbf{Model}} &
\multicolumn{2}{c|}{\textbf{Single}} &
\multicolumn{2}{c|}{\textbf{Interaction}} &
\multicolumn{2}{c}{\textbf{Comparison}} \\
\cline{2-7}
& \textbf{Acc} & \textbf{RMSE} & \textbf{Acc} & \textbf{RMSE} & \textbf{Acc} & \textbf{RMSE} \\
\midrule
\multicolumn{7}{c}{\textit{Proprietary LLMs}} \\
\midrule
GPT-5-5 & 0.772 & 187.438 & 0.777 & 56.096 & 0.789 & 77.750 \\
Opus-4-5 & 0.761 & 105.049 & 0.729 & 51.932 & 0.777 & 71.679 \\
\midrule
\multicolumn{7}{c}{\textit{Open-weight LLMs}} \\
\midrule
Llama-3.1 70B & 0.683 & 84.119 & 0.530 & 38.646 & 0.456 & 64.887 \\
DeepSeek-3-2 & 0.629 & 68.866 & 0.476 & 44.610 & 0.494 & 55.504 \\
\midrule
\multicolumn{7}{c}{\textit{Liquid Foundation Models}} \\
\midrule
2.6B-MMAI, MT & \textbf{0.840} & 58.880 & 0.815 & 27.135 & 0.801 & 48.344 \\
2.6B-MMAI, ST & \textbf{0.840} & 55.954 & \textbf{0.819} & \textbf{25.046} & \textbf{0.810} & 50.598 \\
24B-A2B-MMAI, ST & 0.820 & \textbf{50.250} & 0.780 & 31.080 & 0.790 & \textbf{47.320} \\
\bottomrule
\end{tabular}
}
\caption{\textbf{Accuracy and RMSE on Functional Group Reasoning results on FGBench.}}
\label{tab:fgbench_res}
\end{table}

\paragraph{Functional group reasoning.} On FGBench (Table~\ref{tab:fgbench_res}), the improvements over the base models are broad and systematic across single-group, interaction, and comparison settings, for both boolean and value prediction. This indicates that MMAI Gym improves not only instruction following, but also local structure-property reasoning. The adapted smaller LFM2 models are the strongest systems overall, outperforming frontier LLMs baselines. The dense LFM2-2.6B-MMAI model is as competitive as a larger LFM2-24B-A2B-MMAI MoE model, even in a non-reasoning mode, showcasing the innate fluency in understanding the molecular input and functional groups. We observe that reasoning-enabled inference is generally more helpful on the harder numeric prediction tasks, while non-reasoning inference remains a strong lower-compute option for boolean questions.

\paragraph{Single-step retrosynthesis.} On single-step retrosynthesis (Table~\ref{tab:ssrs_res}), MMAI Gym again yields a large improvement. The base LFM2 model is not competitive on ChemCensor-based plausibility metrics, whereas the adapted models become competitive with top-tier proprietary and chemical generalist LLMs. On URSA-expert-2026, the MMAI-Gym models combine high uniqueness with strong aggregate ChemCensor scores. Unlike baselines that achieve high maximum plausibility for only a narrow set of predictions, they maintain strong average plausibility across multiple unique candidates. Additional results in \autoref{app:ssrs_res} show that enabling reasoning consistently improves retrosynthetic plausibility, indicating that additional inference-time computation supports the identification of more chemically plausible disconnections. Overall, these results indicate that MMAI Gym strengthens procedural chemical reasoning rather than only property prediction.

\begin{table}[t]
\centering
\resizebox{1\linewidth}{!}{
    \begin{tabular}{@{}l|c|cccc@{}}
\toprule
\multirow{2}{*}{\textbf{Model}} &
\multirow{2}{*}{\textbf{Unique}} & \multirow{2}{*}{\textbf{Max}} & \multicolumn{3}{c}{\textbf{Av. PT-Top-K CC}} \\
\cline{4-6}
&  &  & \textbf{@3} & \textbf{@5} & \textbf{@10} \\
\midrule
\multicolumn{6}{c}{\textit{Proprietary LLMs}} \\
\midrule
Gemini 3 Flash preview & 40\% & \textbf{1.80} & 1.21 & 0.83 & 0.42 \\
Grok-4.1 & 43\% & 1.75 & 1.29 & 0.93 & 0.49 \\
\midrule
\multicolumn{6}{c}{\textit{Open-weight Chemical Specialist Models}} \\
\midrule
NatureLM & 27\% & 1.57 & 0.97 & 0.61 & 0.30 \\
C3LM & 42\% & 1.69 & 1.14 & 0.78 & 0.40 \\
\midrule
\multicolumn{6}{c}{\textit{Liquid Foundation Models}} \\
\midrule
2.6B-MMAI, MT & \textbf{94\%} & 1.45 & 0.84 & 0.55 & 0.28 \\
2.6B-MMAI, ST & 93\% & 1.43 & 0.90 & 0.60 & 0.30 \\
24B-A2B-MMAI, ST & 70\% & 1.73 & \textbf{1.38} & \textbf{1.09} & \textbf{0.63} \\
\bottomrule
\end{tabular}
}
\caption{\textbf{Single-Step Retrosynthesis benchmarking on URSA-expert-2026 test set.}}
\label{tab:ssrs_res}
\end{table}

\paragraph{ADMET/PK and toxicity prediction.} On the TDC ADMET/PK and toxicity benchmarks (Table~\ref{tab:tdc_res}), the best MMAI-trained variants are competitive with substantially larger therapeutic language models and surpass them on several tasks, although specialist leaderboard models still retain an advantage on some difficult regression endpoints. Notably, LFM-2.6B-MMAI achieves best-in-class performance among foundation models on lipophilicity, drug-induced liver injury, and blood-brain barrier tasks, while LFM2-24B-MMAI excels on CYP enzyme predictions. Specialist TDC models remain strongest on most endpoints, highlighting the remaining gap to task-specific architectures. Overall, the results indicate that domain adaptation can compensate for substantial differences in model scale: despite being approximately ten times smaller than TxGemma-27B, LFM2-2.6B-MMAI remains competitive with or superior to it across most of the benchmark.

\paragraph{Spatial molecular distribution learning.} On spatial molecular distribution learning (Table~\ref{tab:distribution_learning_3d}), MMAI Gym extends effectively to structured 3D molecular generation. We omit the base LFM2 models from this comparison because they generally fail to follow the required 3D molecule output format, making reliable evaluation infeasible. Among the evaluated systems, the MMAI-trained LFM models are competitive with strong prior baselines, with LFM2-2.6B-MMAI performing particularly well on geometric distribution metrics such as bond lengths, bond angles, dihedrals, and ring-size statistics. At the same time, some baselines remain stronger on parts of the discrete bond and ring distributions, indicating that 3D molecular generation remains a challenging and heterogeneous task.

\begin{table}[t]
\centering
\small
\resizebox{1\linewidth}{!}{
    \begin{tabular}{l|c|cc|c|c}
\toprule
\multirow{2}{*}{\textbf{Task}} & \multirow{2}{*}{\textbf{Metric}} &
\multicolumn{2}{c|}{\textbf{LFM2-MMAI}} & \multirow{2}{*}{\textbf{TxGemma-27B}} &
\multirow{2}{*}{\textbf{TDC SOTA}} \\
& & \textbf{2.6B} & \textbf{24B} & & \\
\midrule
Caco2 Wang              & MAE ($\downarrow$)     & \underline{0.347} & 0.36 & 0.401 & \textbf{0.256} \\
Lipophilicity AZ        & MAE ($\downarrow$)     & \textbf{0.434} & 0.57 & 0.538 & \underline{0.456} \\
Solubility AqSolDB      & MAE ($\downarrow$)     & \underline{0.812} & 0.84 & 0.907 & \textbf{0.741} \\
PPBR AZ                 & MAE ($\downarrow$)     & \underline{8.006} & 8.62 & 9.048 & \textbf{7.440} \\
VDss Lombardo           & Spearman ($\uparrow$)  & 0.591 & \underline{0.63} & 0.559 & \textbf{0.713} \\
Half Life Obach         & Spearman ($\uparrow$)  & 0.381 & \underline{0.46} & 0.458 & \textbf{0.576} \\
Clearance Microsome AZ  & Spearman ($\uparrow$)  & 0.531 & \underline{0.59} & 0.462 & \textbf{0.630} \\
Clearance Hepatocyte AZ & Spearman ($\uparrow$)  & 0.356 & \underline{0.48} & 0.260 & \textbf{0.536} \\
Pgp Broccatelli         & AUROC ($\uparrow$)     & 0.863 & 0.91 & \underline{0.937} & \textbf{0.938} \\
HIA Hou                 & AUROC ($\uparrow$)     & 0.987 & \underline{0.99} & 0.988 & \textbf{0.993} \\
BBB Martins             & AUROC ($\uparrow$)     & \underline{0.924} & \textbf{0.96} & 0.908 & \underline{0.924} \\
Bioavailability Ma      & AUROC ($\uparrow$)     & \underline{0.751} & 0.74 & 0.694 & \textbf{0.942} \\
CYP2C9 Substrate CM     & AUPRC ($\uparrow$)     & \underline{0.444} & 0.4 & 0.438 & \textbf{0.474} \\
CYP2C9 Veith            & AUPRC ($\uparrow$)     & 0.751 & \underline{0.78} & 0.683 & \textbf{0.859} \\
CYP2D6 Substrate CM     & AUPRC ($\uparrow$)     & 0.630 & 0.69 & \underline{0.711} & \textbf{0.736} \\
CYP2D6 Veith            & AUPRC ($\uparrow$)     & 0.585 & 0.61 & \underline{0.683} & \textbf{0.790} \\
CYP3A4 Substrate CM     & AUROC ($\uparrow$)     & 0.681 & \textbf{0.7} & \underline{0.691} & 0.667 \\
CYP3A4 Veith            & AUPRC ($\uparrow$)     & 0.855 & \underline{0.86} & 0.854 & \textbf{0.916} \\
\midrule
LD50 Zhu                & MAE ($\downarrow$)     & 0.703 & 0.67 & \underline{0.627} & \textbf{0.552} \\
DILI                    & AUROC ($\uparrow$)     & \underline{0.937} & 0.89 & 0.886 & \textbf{0.956} \\
hERG                    & AUROC ($\uparrow$)     & 0.787 & 0.83 & \textbf{0.885} & \underline{0.880} \\
AMES                    & AUROC ($\uparrow$)     & 0.782 & 0.81 & \underline{0.816} & \textbf{0.871} \\
\bottomrule
\end{tabular}
}
\caption{\textbf{ADMET Properties Prediction results on TDC.} Best is \textbf{bold}, second best is \underline{underlined}.}
\label{tab:tdc_res}
\end{table}

\paragraph{Pocket-conditioned 3D ligand generation.} On pocket-conditioned 3D ligand generation (Table~\ref{tab:pocket_conditioned}), the MMAI-trained models demonstrate strong performance on structure-based drug design. Most notably, both LFM2-MMAI variants achieve the best UniDock binding affinity scores among all language models, substantially outperforming both proprietary and open-weight LLM baselines and approaching the performance of specialist diffusion models. An interesting contrast emerges with GPT-5.5, which produces highly valid molecules with strong geometric plausibility but only moderate binding affinity. In contrast, the MMAI Gym-trained models optimize more directly for the key objective, generating high-affinity ligands while maintaining competitive validity rates. This suggests MMAI Gym successfully adapts LFM2 to prioritize binding affinity over conservative, geometrically safe but less optimized structures. While specialist diffusion models still achieve the strongest overall binding predictions, the LFM2-MMAI models represent a significant advancement for language model-based structure-based drug design.

\paragraph{Ablation study.}
We further isolate the contributions of the main training and inference choices by comparing the base and MMAI Gym-trained models, multi-task and single-task, with and without reasoning, and the 2.6B and 24B model scales. Overall, MMAI Gym training is the primary source of improvement; multi-task generally provides better robustness, whereas single-task can benefit individual endpoints. Reasoning is most helpful for compositionally demanding tasks, while scaling yields task-dependent rather than uniform gains. Additional details of these comparisons are provided in \autoref{app:ablation}.

\section{Conclusion} \label{sec:conclusion}

\begin{table}[t]
\centering
\small
\resizebox{0.95\linewidth}{!}{
    \begin{tabular}{c|c|c|c|c|c}
\toprule
\multirow{2}{*}{\textbf{Metrics}} & \multicolumn{3}{c|}{\textbf{Baselines}} & \multicolumn{2}{c}{\textbf{LFM-MMAI-Gym}} \\
\cline{2-6}
 & \textbf{MolDiff} & \textbf{OpenLLaMA} & \textbf{nach0} & \textbf{2.6B} & \textbf{24B-A2B} \\
\midrule
JS. bond lengths ($\downarrow$) & 0.436 & 0.257 & 0.148 & \textbf{0.116} & 0.122 \\
JS. bond angles ($\downarrow$) & 0.181 & 0.136 & 0.096 & \textbf{0.094} & 0.098 \\
JS. dihedral angles ($\downarrow$) & 0.198 & 0.110 & 0.110 & \textbf{0.101} & 0.139 \\
\midrule
JS. \# bonds per atoms ($\downarrow$) & 0.121 & \textbf{0.064} & 0.111 & 0.118 & 0.131 \\
JS. freq. bond types ($\downarrow$) & 0.170 & \textbf{0.031} & 0.046 & 0.040 & 0.066 \\
JS. freq. bond pairs ($\downarrow$) & 0.153 & \textbf{0.028} & 0.037 & 0.046 & 0.080 \\
JS. freq. bond triplets ($\downarrow$) & 0.137 & \textbf{0.034} & 0.046 & 0.052 & 0.085 \\
\midrule
JS. \# rings ($\downarrow$) & 0.079 & \textbf{0.033} & 0.56 & 0.081 & 0.111 \\
JS. \# n-sized rings ($\downarrow$) & 0.102 & 0.024 & 0.026 & \textbf{0.019} & 0.025 \\
\# Intersecting rings ($\uparrow$) & 8 & 8 & 8 & \textbf{9} & 8 \\
\bottomrule
\end{tabular}
}
\caption{\textbf{3D molecular distribution learning.}}
\label{tab:distribution_learning_3d}
\end{table}

Our experiments demonstrate that a compact, domain-adapted model can achieve specialist-level performance on several drug-discovery benchmarks without requiring frontier-scale model size. With the right combination of molecular data, training procedures, and efficient architectures, it is possible to train a competitive language model across a wide range of practical tasks. 

Domain-adapted SFT+RFT is substantially more effective than using the few-shot prompted foundation models. Moreover, in a few cases, the multi-task model surpasses strong single-task specialists, suggesting that broader training can improve accuracy and robustness without diluting performance.

We observe an even stronger pattern in molecular optimization and functional group reasoning, where the model achieves the highest success rates overall. These results indicate that the model learns actionable structure–property relationships that translate into effective molecular edits. Even in harder tasks, such as single-step retrosynthesis, the smaller adapted model remains competitive and produces results close to substantially larger foundation models, showing that chemistry-focused post-training can compensate scale differences in this specialization-versus-few-shot comparison.

While our models do not yet uniformly outperform specialist systems on every benchmark, when multi-task training alone falls short of state-of-the-art performance, category-focused training can narrow the gap without sacrificing efficiency.

All these findings support the main message of this paper. MMAI Gym for Science provides a recipe that can substantially improve the capabilities of foundation models, reliably transforming a base model into a much stronger and more versatile molecular assistant. The resulting models achieve new or near state-of-the-art results and offer strong accuracy-efficiency trade-offs relative to specialists and few-shot-prompted foundation models.

\begin{table}[t]
\centering
\small
\resizebox{1\linewidth}{!}{
    \begin{tabular}{l|cc|cc|cc}
\toprule
\multirow{2}{*}{\textbf{Model}} & \multicolumn{4}{c|}{\textbf{Validity}} & \multicolumn{2}{c}{\textbf{Binding}} \\
\cline{2-7}
 & \multicolumn{2}{c|}{PB Intra} & \multicolumn{2}{c|}{Int. Energy} & \multicolumn{2}{c}{UniDock} \\
 & \textbf{CD} & \textbf{PL} & \textbf{CD} & \textbf{PL} & \textbf{CD} & \textbf{PL} \\
\midrule
PocketXMol & \textbf{95\%} & \textbf{97\%} & \textbf{99\%} & \textbf{99\%} & -8.1 & -7.0 \\
MolSnapper & 89\% & 85\% & 83\% & 80\% & \textbf{-8.6} & \textbf{-8.2} \\
\midrule
GPT-5.5 & \textbf{95\%} & 95\% & 96\% & 96\% & -4.7 & -4.4 \\
Gemini 3.1-Pro & 86\% & 87\% & 96\% & 97\% & -5.2 & -4.7 \\
\midrule
LFM2-2.6B-MMAI & 56\% & 52\% & 79\% & 83\% & -7.3 & -7.0 \\
LFM2-24B-A2B-MMAI & 55\% & 57\% & 85\% & 81\% & -7.4 & -7.0 \\
\bottomrule
\end{tabular}
}
\caption{\textbf{Pocket-conditioned 3D generation.}}
\label{tab:pocket_conditioned}
\end{table}

\section{Ethical Considerations and Limitations} 
\label{sec:limitations}

While MMAI Gym training substantially improves molecular AI capabilities across diverse tasks, several limitations remain in this version, which merit discussion and suggest directions for future work.

\paragraph{Safety and dual-use risks.} Molecular language models could be misused to generate harmful, toxic, or dual-use compounds. Although MMAI Gym focuses on drug-like and therapeutically relevant chemical space, this does not eliminate misuse risk. In particular, toxicity-related tasks are designed for property prediction rather than toxicity maximization, but their objectives could be inverted to guide harmful optimization. The MMAI-Gym models have not undergone adversarial red-teaming or safety alignment, and therefore should not be deployed without appropriate safeguards.

\paragraph{Release policy and access controls.} We do not plan to release the LFM2-MMAI checkpoints or the complete MMAI Gym training environment for unrestricted public download. Model access may be requested through the \href{https://www.liquid.ai/request-access-mmai}{LFM2-MMAI access form} and is granted only after review; requests may be denied and access revoked if misuse is suspected. We intend to release non-sensitive benchmark definitions and evaluation resources, but not the processed task prompts, reasoning traces, in-house data, or tooling that could materially lower the barrier to reproducing or adapting the system for harmful objectives. This policy applies even though many underlying source datasets are already publicly available, as the curated tasks and reasoning data provide capability beyond the original resources. Before expanding access, we plan to investigate safeguards such as refusal training, chemical-structure screening, input and output filtering, rate limits, and misuse monitoring. Until these measures are validated, application review, restricted distribution, and withholding capability-enhancing training artifacts are our primary safeguards.

\paragraph{Evaluation limitations.} Our evaluation relies primarily on established computational benchmarks, which may not fully capture the complexity and diversity of real-world drug discovery workflows. Benchmarks can contain distributional artifacts, have limited coverage of edge cases, and success on them does not guarantee performance on proprietary industrial datasets or novel chemical spaces outside the training distribution. Furthermore, our work does not include prospective experimental validation of generated molecules through synthesis and biological assays, relying solely on already existing public ground truth values in this work. The true utility of model-generated molecules and predictions on novel out-of-distribution molecules can only be established through wet-lab validation, which remains a critical next step for translation to practical drug discovery applications.

\paragraph{Performance gaps with specialist systems.} Although the MMAI-trained LFM2 models achieve strong performance across most benchmarks and start to achieve SOTA-level performance on a few, they do not uniformly surpass specialist systems on every metric. Specialized non-LLM architectures trained exclusively on narrow tasks maintain advantages in specific domains: for example, dedicated regression models on certain TDC endpoints (e.g., Clearance, VDss) and diffusion-based generators on some 3D molecular generation metrics. This indicates that domain-specific architectural inductive biases and task-focused optimization remain competitive with general-purpose language models when evaluated on the precise tasks they were designed to solve.

\paragraph{Geometric constraint satisfaction.} On pocket-conditioned generation, the MMAI-trained models achieve excellent binding affinity predictions but struggle with fine-grained geometric constraints. While strong docking scores indicate good predicted binding affinity, the failure to satisfy all geometric constraints suggests the models may generate structures that are chemically plausible but geometrically imperfect. We attribute much of this gap to data: high-fidelity pocket-based 3D generation likely requires more specialized training signal than public datasets provide, including fragment-level placement within a pocket and a broader range of 3D conditioning modalities. More work and benchmarks are needed to showcase geometric fit in 3D generation with multiple conditions.

\paragraph{Domain and task specialization.} By design, MMAI Gym adapts a general-purpose base model into a multi-task specialist but only for drug discovery, so the resulting models are tuned to the molecular domains and task families represented in our training mixture. Their strengths are therefore concentrated within this scope: performance is most reliable on tasks that resemble those seen during training, while extrapolation to substantially different task formulations, instructions, or problem framings outside the trained distribution is more limited. Likewise, reasoning over out-of-distribution inputs-novel task types, unfamiliar prompt formats, or problems requiring capabilities beyond the curated molecular curriculum-tends to be comparatively weaker, and the strong domain adaptation can temper some of the broad, general-purpose abilities of the base model. We regard this specialization as a deliberate trade-off rather than a defect, but it does constrain plug-and-play use beyond drug discovery. Future work includes broadening task and domain coverage, retaining general reasoning alongside domain expertise, and explicitly training for instruction-level generalization so the models extrapolate better to unseen tasks.

\newpage

\bibliography{iclr2026_conference}

\newpage
\appendix

\section{Datasets and Training Corpus}
\label{app:datasets}

\paragraph{Corpus scope and definition.}
MMAI Gym for Science comprises tasks across six high-level categories: 2D molecular tasks, 3D molecular tasks, 2D protein tasks, 3D protein tasks, drug-gene interaction tasks, and cross-domain tasks that jointly use multiple modalities. These categories include regression, classification, and generation problems.

\paragraph{Molecular data.}
The 2D molecular tasks draw on TDC~\citep{huang2021tdc}, MOSES~\citep{polykovskiy2020moses}, FGBench~\citep{liu2025fgbench}, MuMO-Instruct~\citep{dey2025mumoinstruct}, CREED~\citep{zagribelnyy2026chemcensor}, and MolTextNet~\citep{zhu2025moltextnet}. These resources support tasks such as ADME and toxicity prediction, functional-group reasoning, multi-objective molecular optimization, unconditional molecular generation, retrosynthesis, and molecule-text modeling. The TDC component contains multiple individual ADME, pharmacokinetic, and toxicity tasks, which are listed separately in Table~\ref{tab:dataset_statistics}.

The 3D molecular tasks use GEOM~\citep{axelrod2022geom}, ZINC~\citep{irwin2005zinc}, and QMUGS~\citep{isert2022qmugs}. These datasets provide molecular conformations for learning molecular geometry, 3D pose generation, and spatial molecular distributions.

\paragraph{Protein and interaction data.}
The protein tasks use ProteinLMBench~\citep{shen2024proteinlmbench}, the AlphaFold Protein Structure Database~\citep{varadi2023alphafolddb} and the BC40 subset of PDB~\citep{berman2000pdb}. Together, these resources support tasks involving protein sequences, protein structures, and peptide-protein complexes.

Interaction tasks are constructed from  STRING~\citep{szklarczyk2025stringdb}.

\paragraph{Cross-domain and in-house data.}
Tasks combining molecular, protein, structural, and natural-language modalities incorporate CrossDocked2020~\citep{francoeur2020crossdocked}, L+M-24~\citep{edwards2024lm24}. CrossDocked2020 supports pocket-conditioned molecular generation, whereas the remaining resources provide language-molecule and instruction-following examples. Public datasets in each major category are complemented by high-quality in-house data.

\paragraph{Processing and serialization.}
We convert each source into a unified instruction-following representation while retaining the inputs and outputs required by its underlying task.

Property-prediction examples are serialized as regression or classification questions. Molecular-optimization examples pair an initial molecule with requested property changes, while retrosynthesis map a target product to plausible reactants.  

Molecular entities may be represented using SMILES, SELFIES, or IUPAC names, with randomized format conversion and non-canonical traversal used as data augmentation. Protein sequences are represented using FASTA format.

For 3D tasks, molecular graphs or protein residues are serialized together with their atomic coordinates following the representations described in \autoref{sec:mmai_gym}. 

Dataset-specific validation, filtering, and format
normalization are applied before tokenization. The evaluation pipeline also uses public and internal out-of-distribution benchmarks to assess generalization to novel chemical and biological settings.

\paragraph{Data balancing.}
Training data vary substantially in size across datasets and modalities. To prevent large sources from dominating multi-task training, we use hierarchical sampling. We first sample uniformly from the six high-level task categories, then uniformly select a dataset-task pair within that category, and finally sample an object from the selected pair. Thus, the raw object counts in Table~\ref{tab:dataset_statistics} describe corpus composition but do not directly determine the frequency with which each task is observed during training.

\paragraph{Dataset statistics.}
Table~\ref{tab:dataset_statistics} reports the numbers of processed objects and their mean token lengths. 

\begin{figure*}[!ht]
    \vspace{-2.5em}
    \centering
    \includegraphics[width=0.85\linewidth]{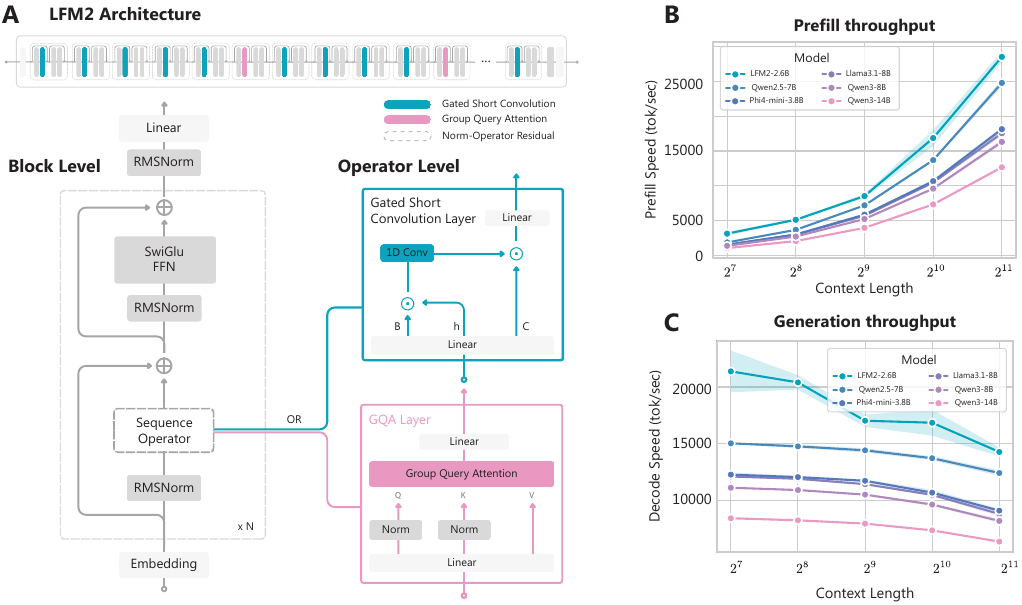}
    \vspace{-1em}
    \caption{\textbf{LFM2 Architecture \& Efficiency}. (a) LFM2 is a hybrid language model comprised mainly of gated short convolution sequence operators offering improved throughput and efficency \citep{amini2025lfm2}. (b,c) LFM2 prefill processing speed and decode speed with increasing prefill. }
    \label{fig:arch}
    \vspace{-0.6em}
\end{figure*}

\section{Liquid Foundation Model}
\label{app:lfm}

Full softmax attention \citep{vaswani2017transformer} based language models incur substantial computational and memory costs, particularly for long-context and resource constrained deployment scenarios. These limitations motivate the need for efficient alternatives, including hybrid architectures in which the majority of layers use sub-quadratic sequence operators, such as linear attention or state space models (SSMs) \citep{dao2024mamba2,yang2024gated,katharopoulos2020transformers}, while a smaller number of softmax attention layers are interleaved to serve as global token mixers. This design has been shown to achieve a favorable Pareto tradeoff between model expressivity and efficiency \citep{kimiteam2025kimik2openagentic,blakeman2025nemotron}.

To this end, we utilize LFM2-2.6B \citep{amini2025lfm2}, a 2.6B parameter autoregressive hybrid language model from the Liquid Foundation Model (LFM) family, specifically designed for long context inference under latency and memory constraints. LFM2-2.6B (Figure~\ref{fig:arch}a) is composed of two sequence operator primitives: primarily \textit{gated short convolutions} (ShortConv) for sequence featurization and a few \textit{grouped-query softmax attention} (GQA) \citep{ainslie2023gqa} for global sequence mixing. 

Each LFM2 block is composed of two sequential \texttt{Norm-Operator-Residual} schemes: first, where the sequence operator is either a ShortConv or GQA, and second, where the operator is SwiGLU feed-forward layers \citep{shazeer2020glu} for the position-wise MLP readout. Each block utilizes a pre-Root Mean Squared Normalization (pre-RMSNorm) \citep{zhang2019root,jiang2023pre}, which omits the centering step in LayerNorms for improved efficiency.

The gated short convolution operator, for a hidden sequence $h \in \mathbb{R}^{L \times d}$, computes the following,
\begin{align}
& (B, C, \Tilde{h})  = hW_{\text{in}}\\
& o  = \left(C \odot \text{Conv}_k(B \odot \Tilde{h})\right) W_\text{out} 
\end{align}
where $W_\text{in} \in \mathbb{R}^{d \times 3d}, W_\text{out} \in \mathbb{R}^{d \times d}$ are learned linear projections, Conv$_k$ denotes a depthwise 1D convolution with $k=3$, and $\odot$ denotes the canonical Hadamard product. These local operators encode short range ``motifs" in the sequence, which are subsequently integrated over long contexts by the small number of global attention operators. The global mixing layers use GQA \citep{ainslie2023gqa}, which partitions queries into groups that share keys and values, reducing overhead.

Empirically, these architectural design choices translate into substantial efficiency gains during both prefill and decode. Across a range of prefill context lengths, LFM2-2.6B consistently achieves higher throughput than the other smallest open-weight baseline models that were also finetuned using MMAI Gym, as well as Phi-4-mini \citep{Abouelenin2025Phi4MiniTR}, another model optimized for compute, memory, and latency bound environments (Figure~\ref{fig:arch}b,c). This trend reflects the contrast between the LFM hybrid design, with its linear-time ShortConv operators and limited GQA, and attention-dominant transformers that incur quadratic scaling at longer contexts.

\section{Training Protocol}
\label{app:training}

\paragraph{SFT:} At each optimization step of $100,000$, we process an aggregate context of $3,145,728$ tokens across all GPUs, using sequence packing \cite{brown2020packing} to fill each device’s available context window with multiple training examples. We conduct supervised fine-tuning on the MMAI Gym corpus with the \texttt{AdamW} optimizer, using a learning rate of $1 \text{x} 10^{-5}$, linear warmup for the first 5\% of the training steps, and a cosine decay schedule for the remainder. 

\paragraph{RFT:} Online RFT was performed using Group Relative Policy Optimization (GRPO, \cite{shao2024deepseekmathpushinglimitsmathematical}), using the same tasks and prompts as in SFT. We use a group-size of $8$, with $16$ groups per global batch, resulting in a batch-size of $128$. With a maximum sequence length of $16,384$, this results in a maximum of $2,097,152$ tokens processed per step across all GPUs (we use one update per batch). However, the effective token throughput due to variable completion lengths was approximately $130,000$ tokens per step. Due to the significant number of SFT steps to fully learn molecular languages, we use a high sampling temperature of $1.4$ to encourage generation diversity. Also, due to the diversity of tasks and task-specific reward functions, we keep KL-regularization with $\beta=0.4$. Concerning the optimizer, we use AdamW with a constant learning rate of $1 \text{x} 10^{-6}$ and weight-decay of $0.05$. Finally, we use Deepspeed stage 3 \citep{rasley2020deepspeed}, with gradient checkpointing and BF16 precision.

To explore the benefits of RFT, we either perform RFT on all tasks concurrently to train an RFT "generalist" model, or on a single task for a "specialist" model. For the generalist case, we sample prompts uniformly across all tasks, and use one task-specific reward function according to the sampled prompt (plus formatting rewards). Due to the high wall-time required to compute rewards for molecular generation tasks, all RFT setups involving generation were trained over $1,000$ steps. 

\section{Task-specific Rewards}
\label{app:rl_rewards}

MMAI Gym includes 2 generic reward functions applicable to all tasks.

\paragraph{Format} The \emph{format} reward ensures the reasoning is properly enclosed in \tok{think} tags. Formally, $r_\text{format} = -1 + \mathbbm{1}_{n_{\text{\tok{think}}}=1} + \mathbbm{1}_{n_{\text{\tok{\textbackslash think}}}=1}$, where $n_{\text{\tok{think}}}$ counts the number of occurrences of \tok{think}, and $\mathbbm{1}$ is the indicator function. 

\paragraph{Thinking length} The \emph{thinking} reward encourages longer thinking/reasoning sequences. Among the large number of task-datasets present in MMAI Gym, some do not incorporate reasoning, resulting in sequences where the answer follows immediately after the question. This results in a behavior where the model sometimes output an answer immediately after a question, which hampers the sequences generated during reinforcement learning post-training. To mitigate this, we encourage the model to generate longer sequences, up to a saturation point of $5,000$ characters, i.e., $r_\text{think} = \min(1, -1 + |o|/ {2500})$.\newline

\noindent In addition to these rewards, each task perceives its own task-specific reward.

\paragraph{Classification.} All considered classification tasks are question-answering problems, where the model must choose between a set of possible answers. Given the ground-truth $g$ and answer $a$, $r_{\text{qa}} = \mathbbm{1}_{||a|-|g||<3} + \mathbbm{1}_{a=g} -1$. The first part constraints the difference in length between the ground-truth and prediction to be less than $3$ characters, ensuring that the model only outputs the provided choice (e.g., "True") without any additional justification. 

\paragraph{Regression.} For regression tasks, we define the reward as $r_{\text{reg}} = \mathbbm{1}_{||a|-|g||<3} - \frac{|a-g|}{(\max(\mathcal{A})-\min(\mathcal{A})}$, where $\mathcal{A}$ is the dataset of all the answers, and the second part corresponds to the normalized absolute value between the answer and the ground-truth. Analogously to $r_\text{qa}$, the first part encourages the predicted answer to have a similar format to the ground-truth (in this case, similar precision).

\paragraph{Molecular generation.}  Finally, molecular generation tasks verify that the generated answer is a valid molecule as determined by RDKit \cite{landrum2023rdkit}, i.e. $r_\text{gen\_only} = 2*\mathbbm{1}_{\text{is\_valid(a)}}-1$. Alternatively, if there exists a set of verified ground truth values $G$ for a given task, then presence in the ground-truth set supersedes the validity. The reward for these tasks is  $r_\text{gen\_gt} = 2*\mathbbm{1}_{a \in G}-1$.

\section{Extended Experiments}
\label{app:ext_experiments}

In all benchmarks where possible and applicable, we report SFT+Multi-task RFT (\textbf{MT}) and SFT+Single-Task RFT (\textbf{ST}) LFM2-MMAI models with enabled (\texttt{/think}) and disabled(\texttt{/no\_think}) reasoning on inference.

\begin{table*}[t]
\centering
\resizebox{1\textwidth}{!}{
    \begin{tabular}{l|ccc|ccc|ccc|ccc|ccc}
\toprule
\multirow{2}{*}{\textbf{Model}} &
\multicolumn{3}{c|}{\textbf{BDP}} &
\multicolumn{3}{c|}{\textbf{BDQ}} &
\multicolumn{3}{c|}{\textbf{BPQ}} &
\multicolumn{3}{c|}{\textbf{DPQ}} &
\multicolumn{3}{c}{\textbf{BDPQ}} \\
& SR  & Sim  & RI
& SR  & Sim  & RI
& SR  & Sim  & RI
& SR  & Sim  & RI
& SR  & Sim  & RI  \\
\midrule
\multicolumn{16}{c}{\textit{Proprietary LLMs}} \\
\midrule
Grok 4.1
& 7.2 & \textbf{0.71} & 1.95
& 3.8 & \underline{0.62} & 1.76
& 66.6 & \underline{0.64} & 0.87
& 14.6 & 0.50 & 2.20
& 13.2 & 0.51 & 2.77 \\
Gemini-3.1-pro
& 53.4 & 0.44 & 4.26
& 58.4 & 0.45 & 6.03
& 75.0 & 0.52 & 1.40
& 33.6 & 0.41 & 3.46
& 31.4 & 0.34 & 5.31 \\
GPT-5.2
& 39.0 & \underline{0.68} & 1.27
& 49.2 & \underline{0.62} & 3.16
& 85.4 & \textbf{0.67} & 0.83
& 28.6 & \textbf{0.57} & 0.64
& 18.6 & 0.59 & 1.01 \\
GPT-5.4
& 68.6 & 0.57 & 2.27
& 85.6 & 0.51 & 6.32
& 94.0 & 0.57 & 1.16
& 50.4 & 0.48 & 1.91
& 38.0 & 0.5 & 2.34 \\
GPT-5.5 
& 83.2 & 0.53 & 2.47 
& \textbf{91.6} & 0.45 & \textbf{7.62}
& 92.6 & 0.52 & 1.43 
& \underline{72.4} & 0.45 & 3.44 
& \textbf{77.0} & 0.36 & 4.57 \\ 
Claude 4.5 Sonnet
& 71.4 & 0.52 & 2.73
& 55.0 & 0.37 & 3.83
& 97.4 & 0.46 & 1.54
& 55.2 & 0.37 & 2.55
& 32.2 & 0.38 & 2.61 \\
Claude 4.6 Opus
& 69.4 & 0.42 & 3.87
& 62.2 & 0.43 & \underline{6.58}
& \underline{98.6} & 0.45 & 1.69
& \textbf{73.2} & 0.30 & \textbf{6.48}
& 56.4 & 0.32 & 4.82 \\
\midrule
\multicolumn{16}{c}{\textit{Foundational LLMs for Chemistry}} \\
\midrule
ChemLLM
& 0.2 & 0.17 & 1.2
& 1.0 & 0.55 & 0.82
& 4.8 & 0.29 & 0.96
& 0.6 & 0.28 & 0.42
& 0.0 & n/a & n/a \\
$\mathrm{LlaSMol}_{\mathrm{Mistral}}$
& 43.6 & 0.62 & 1.09
& 31.4 & \textbf{0.66} & 0.93
& 86 & 0.58 & 0.84
& 24 & \textbf{0.57} & 0.61
& 14 & \underline{0.62} & 1.03 \\
\midrule
\multicolumn{16}{c}{\textit{Task-specific non-LLMs}} \\
\midrule
Prompt-MolOpt
& 12.2 & 0.12 & \textbf{7.46}
& 23.2 & 0.10 & 5.4
& 15.8 & 0.10 & 1.5
& 23.6 & 0.10 & 5.46
& 6.6 & 0.11 & \underline{5.36} \\
\midrule
\multicolumn{16}{c}{\textit{Task-specific LLMs}} \\
\midrule
$\mathrm{GeLLM}^{3}\,\mathrm{O}\text{-}3_{\mathrm{Mistral}}$
& 84.8 & 0.47 & 4.3
& 87.0 & 0.47 & 5.61
& 93.0 & 0.46 & 1.49
& 62.8 & 0.37 & 3.87
& -- & -- & -- \\
$\mathrm{GeLLM}^{3}\,\mathrm{O}\text{-}3_{\mathrm{Llama}}$
& \textbf{86.8} & 0.48 & 4.38
& 90.0 & 0.46 & 5.66
& 94.0 & 0.5 & 1.38
& 60.6 & 0.44 & 3.76
& -- & -- & -- \\
$\mathrm{GeLLM}^{3}\,\mathrm{O}\text{-}4_{\mathrm{Mistral}}$
& 71.6 & 0.49 & 3.27
& 57.4 & 0.55 & 2.56
& 90.2 & 0.46 & 1.41
& 54.0 & 0.44 & 3.02
& 30.0 & 0.48 & 3.44 \\
$\mathrm{GeLLM}^{3}\,\mathrm{O}\text{-}4_{\mathrm{Llama}}$
& 53.6 & 0.63 & 1.94
& 48.6 & 0.59 & 1.29
& 93.4 & 0.59 & 1.12
& 39.6 & \textbf{0.57} & 1.32
& 28.0 & \textbf{0.66} & 1.02 \\
\midrule
\multicolumn{16}{c}{\textit{Generalist LLMs}} \\
\midrule
$\mathrm{GeLLM}^{3}\,\mathrm{O}\text{-}\mathrm{P}(4)_{\mathrm{Mistral}}$
& 81.4 & 0.55 & 3.95
& 82.6 & 0.56 & 5.24
& 96.2 & 0.52 & 1.52
& 66.6 & 0.53 & 2.41
& 57.4 & 0.52 & 3.04 \\
$\mathrm{GeLLM}^{3}\,\mathrm{O}\text{-}\mathrm{P}(4)_{\mathrm{Llama}}$
& 80.4 & 0.54 & 3.6
& 81.4 & 0.56 & 4.81
& 93.8 & 0.47 & 1.64
& 61.4 & 0.5 & 2.02
& 49.8 & 0.48 & 3.26 \\
$\mathrm{GeLLM}^{3}\,\mathrm{O}\text{-}\mathrm{P}(6)_{\mathrm{Mistral}}$
& 83.0 & 0.57 & 3.6
& 85.8 & \underline{0.59} & 4.78
& 96.8 & 0.53 & 1.48
& 60.8 & \underline{0.54} & 2.16
& 54.0 & 0.54 & 3.09 \\
$\mathrm{GeLLM}^{3}\,\mathrm{O}\text{-}\mathrm{P}(6)_{\mathrm{Llama}}$
& 77.0 & 0.53 & 3.73
& 79.6 & 0.56 & 5.05
& 95.0 & 0.47 & 1.66
& 57.0 & 0.49 & 2.50
& 52.2 & 0.49 & 3.48 \\
\midrule
\multicolumn{16}{c}{\textit{Liquid Foundation Models}} \\
\midrule
LFM2-2.6B
& 38.0 & 0.18 & 1.16
& 31.4 & 0.08 & 0.88
& 80.2 & 0.19 & \textbf{2.09}
& 32 & 0.11 & 0.76
& 17.4 & 0.11 & 1.62 \\
LFM2-2.6B-MMAI, MT
& \underline{86.0} & 0.42 & 4.09
& \textbf{91.6} & 0.50 & 4.54
& \textbf{98.8} & 0.41 & 1.82  
& 71.2 & 0.40 & 2.72
& 56.4 & 0.38 & 3.67 \\
LFM2-2.6B-MMAI, ST
& \underline{86.0} & 0.41 & 4.35
& \underline{91.2} & 0.51 & 4.42
& \textbf{98.8} & 0.41 & 1.88
& 69.8 & 0.39 & 2.95
& 57.8 & 0.37 & 4.00 \\
\midrule
LFM2-24B-A2B
& 9.6 & 0.39 & \underline{5.04}
& 13.6 & 0.34 & 5.46
& 39.0 & 0.39 & 1.46
& 11.4 & 0.26 & \underline{6.04}
& 11.4 & 0.21 & \textbf{6.95} \\
LFM2-24B-A2B-MMAI, ST
& 68.2 & 0.42 & 3.50
& 78.8 & 0.46 & 4.39
& 97.2 & 0.37 & \underline{2.00}
& 60.2 & 0.36 & 2.83
& \underline{60.4} & 0.33 & 3.24 \\
\bottomrule
\end{tabular}  
}
\caption{\textbf{Molecular Optimization results on MuMO-Instruct.} We report Success Rate (SR), Similarity (Sim), and Relative Improvement (RI) on five tasks. Best is \textbf{bold}, second best is \underline{underlined}.}
\label{tab:mumo_res_ext}
\end{table*}

\begin{table*}[t]
\centering
\resizebox{1\textwidth}{!}{
    \begin{tabular}{l|cc|cc|cc|cc|cc|cc}
\hline
\multirow{3}{*}{\textbf{Model}} &
\multicolumn{4}{c|}{\textbf{Single}} &
\multicolumn{4}{c|}{\textbf{Interaction}} &
\multicolumn{4}{c}{\textbf{Comparison}} \\
\cline{2-13}
&
\multicolumn{2}{c|}{\textbf{Boolean}} &
\multicolumn{2}{c|}{\textbf{Value}} &
\multicolumn{2}{c|}{\textbf{Boolean}} &
\multicolumn{2}{c|}{\textbf{Value}} &
\multicolumn{2}{c|}{\textbf{Boolean}} &
\multicolumn{2}{c}{\textbf{Value}} \\
\cline{2-13}
&
\textbf{Acc} & \textbf{Val} &
\textbf{RMSE} & \textbf{Val} &
\textbf{Acc} & \textbf{Val} &
\textbf{RMSE} & \textbf{Val} &
\textbf{Acc} & \textbf{Val} &
\textbf{RMSE} & \textbf{Val} \\
\hline
\multicolumn{13}{c}{\textit{Proprietary LLMs}} \\
\midrule
GPT-5-1                       & 0.687 &  \textbf{1.000} & 88.055 &  \textbf{1.000} & 0.526 &  \textbf{1.000} & 41.062 &  \textbf{1.000} & 0.693 &  \textbf{1.000} & 63.233 &  \textbf{1.000} \\
GPT-5-2                       & 0.741 &  \textbf{1.000} & 88.942 &  \textbf{1.000} & 0.531 &  \textbf{1.000} & 42.624 &  \textbf{1.000} & 0.637 &  \textbf{1.000} & 66.223 &  \textbf{1.000} \\
GPT-5-5                       & 0.772 &  \textbf{1.000} & 187.438 &  \textbf{1.000} & 0.777 &  \textbf{1.000} & 56.096 &  \textbf{1.000} & 0.789 &  \textbf{1.000} & 77.750 &  \textbf{1.000} \\
Sonnet-4-5                    & 0.663 &  \textbf{1.000} & 92.274 &  \textbf{1.000} & 0.606 & 0.999 & 45.147 &  \textbf{1.000} & 0.741 &  \textbf{1.000} & 64.468 &  \textbf{1.000} \\
Opus-4-5                      & 0.761 &  \textbf{1.000} & 105.049 &  \textbf{1.000} & 0.729 &  \textbf{1.000} & 51.932 &  \textbf{1.000} & 0.777 &  \textbf{1.000} & 71.679 &  \textbf{1.000} \\
\midrule
\multicolumn{13}{c}{\textit{Open-weight LLMs}} \\
\midrule
DeepSeek-3-2                  & 0.629 & 0.963 & 68.866 & 0.424 & 0.476 & 0.896 & 44.610 & 0.260 & 0.494 & 0.994 & 55.504 & 0.643 \\
Llama-3.1 8B                  & 0.548 & 0.993 & 162.351 & 0.840 & 0.547 & 0.982 & 421.325 & 0.780 & 0.474 & 0.991 & 80.566 & 0.761 \\
Llama-3.1 70B                 & 0.683 &  \textbf{1.000} & 84.119 & 0.973 & 0.530 &  \textbf{1.000} & 38.646 & 0.977 & 0.456 &  \textbf{1.000} & 64.887 & 0.943 \\
Qwen2.5-7B                    & 0.590 & 0.999 & 63.511 & 0.576 & 0.396 & 0.999 & 36.307 & 0.683 & 0.664 &  \textbf{1.000} & 65.471 & 0.223 \\
\midrule
\multicolumn{13}{c}{\textit{Foundational LLMs for Chemistry}} \\
\midrule
ChemLLM-7B                    & 0.233 & 0.997 & 209.584 & 0.629 & 0.235 & 0.997 & 162.742 & 0.566 & 0.250 &  \textbf{1.000} & 65.428 & 0.514 \\
Llama-3-8B-MolInst            & 0.107 & 0.203 & 328.935 & 0.496 & 0.059 & 0.149 & 188.376 & 0.486 & 0.469 & 0.873 & 138.654 & 0.837 \\
LlaSMol-Mistral-7B            & 0.387 & 0.922 & 266.720 & 0.923 & 0.298 & 0.968 & 262.550 & 0.983 & 0.239 &  \textbf{1.000} & 245.298 & 0.924 \\
\midrule
\multicolumn{13}{c}{\textit{Liquid Foundation Models}} \\
\midrule
LFM2-2.6B & 0.740 &  \textbf{1.000} & 69.368 &  \textbf{1.000} & 0.724 &  \textbf{1.000} & 38.905 &  \textbf{1.000} & 0.712 &  \textbf{1.000} & 61.556 &  \textbf{1.000} \\
LFM2-2.6B-MMAI, MT, \texttt{/no\_think}       & 0.838 &  \textbf{1.000} & 62.766 &  \textbf{1.000} & 0.808 &  \textbf{1.000} & 31.018 &  \textbf{1.000} & 0.789 &  \textbf{1.000} & 53.127 &  \textbf{1.000} \\
LFM2-2.6B-MMAI, MT, \texttt{/think}       & \underline{0.840} &  \textbf{1.000} & 58.880 &  \textbf{1.000} & \underline{0.815} &  \textbf{1.000} & \underline{27.135} &  \textbf{1.000} & \underline{0.801} &  \textbf{1.000} & \underline{48.344} &  \textbf{1.000} \\
LFM2-2.6B-MMAI, ST, \texttt{/no\_think}       & \textbf{0.841} &  \textbf{1.000} & 62.902 &  \textbf{1.000} & 0.808 &  \textbf{1.000} & 30.248 &  \textbf{1.000} & 0.796 &  \textbf{1.000} & 53.045 &  \textbf{1.000} \\
LFM2-2.6B-MMAI, ST, \texttt{/think}       & \underline{0.840} &  \textbf{1.000} & \underline{55.954} &  \textbf{1.000} & \textbf{0.819} &  \textbf{1.000} & \textbf{25.046} &  \textbf{1.000} & \textbf{0.810} &  \textbf{1.000} & 50.598 &  \textbf{1.000} \\
\midrule
LFM2-24B-A2B       & 0.420 &  \textbf{1.000} & 69.980 &  \textbf{1.000} & 0.400 &  \textbf{1.000} & 39.520 &  \textbf{1.000} & 0.250 &  \textbf{1.000} & 58.450 &  \textbf{1.000} \\
LFM2-24B-A2B-MMAI, ST, \texttt{/think}       & 0.820 &  \textbf{1.000} & \textbf{50.250} &  \textbf{1.000} & 0.780 &  \textbf{1.000} & 31.080 &  \textbf{1.000} & 0.79 &  \textbf{1.000} & \textbf{47.320} &  \textbf{1.000} \\
\hline
\end{tabular}  
}
\caption{\textbf{Functional Group Reasoning results on FGBench.} We report accuracy \textbf{Acc} for questions, \textbf{RMSE} for regression, and fraction of valid predictions (\textbf{Val}). Best is \textbf{bold}, second best is \underline{underlined}.}
\label{tab:fgbench_res_ext}
\end{table*}

\begin{table*}[t]
\centering
\resizebox{1\textwidth}{!}{
    \begin{tabular}{@{}l|c|cccc|c|cccc@{}}
\toprule
\multirow{3}{*}{\textbf{Model}} &
\multicolumn{5}{c|}{\textbf{URSA-expert-2026}} &
\multicolumn{5}{c}{\textbf{USPTO-50K-test}} \\
\cline{2-6}\cline{7-11}
& \multirow{2}{*}{\textbf{Unique}} & \multirow{2}{*}{\textbf{Max}} & \multicolumn{3}{c|}{\textbf{Av. PT-Top-K CC}}
& \multirow{2}{*}{\textbf{Unique}} & \multirow{2}{*}{\textbf{Max}} & \multicolumn{3}{c}{\textbf{Av. PT-Top-K CC}} \\
&  &  & \textbf{@3} & \textbf{@5} & \textbf{@10}
&  &  & \textbf{@3} & \textbf{@5} & \textbf{@10} \\
\midrule
\multicolumn{11}{c}{\textit{Proprietary LLMs}} \\
\midrule
Grok-4.1                   & 43\% & 1.75 & \underline{1.29} & \underline{0.93} & \underline{0.49} & --   & --   & --   & --   & --   \\
Gemini 2.5 Flash           & 50\% & 0.66 & 0.30 & 0.18 & 0.09 & --   & --   & --   & --   & --   \\
Gemini 3 Flash preview     & 40\% & \textbf{1.80} & 1.21 & 0.83 & 0.42 & --   & --   & --   & --   & --   \\
GPT 5.1                    & 62\% & 0.73 & 0.35 & 0.22 & 0.11 & 59\% & 1.54 & 0.72 & 0.45 & 0.22 \\
GPT 5.2                    & 51\% & 0.85 & 0.47 & 0.29 & 0.15 & 52\% & 2.04 & 1.01 & 0.63 & 0.32 \\
Claude 4.5 Sonnet          & 70\% & 1.45 & 0.88 & 0.58 & 0.29 & 56\% & 3.37 & 1.81 & 1.15 & 0.58 \\
Claude 4.5 Opus            & 54\% & 1.34 & 0.73 & 0.46 & 0.23 & 44\% & 3.31 & 1.62 & 1.00 & 0.50 \\
\midrule
\multicolumn{11}{c}{\textit{Open-weight LLMs}} \\
\midrule
DeepSeek 3.2               & 52\% & 0.39 & 0.16 & 0.10 & 0.05 & 55\% & 1.14 & 0.46 & 0.27 & 0.14 \\
Qwen3 8B \                 & 52\% & 0.00 & 0.00 & 0.00 & 0.00 & 55\% & 0.04 & 0.03 & 0.02 & 0.01 \\
Qwen3 14B \                & 52\% & 0.01 & 0.00 & 0.00 & 0.00 & 54\% & 0.10 & 0.03 & 0.02 & 0.01 \\
Kimi K2                    & 43\% & 1.12 & 0.62 & 0.38 & 0.19 & --   & --   & --   & --   & --   \\
\midrule
\multicolumn{11}{c}{\textit{Open-weight Chemical Specialist Models}} \\
\midrule
ether0                     & 20\% & 1.01 & 0.55 & 0.35 & 0.17 & 19\% & 2.08 & 1.07 & 0.67 & 0.34 \\
NatureLM                   & 27\% & 1.57 & 0.97 & 0.61 & 0.30 & 20\% & 3.99 & 1.85 & 1.14 & 0.57 \\
RetroDFM-R                 & 12\% & 1.52 & 0.70 & 0.43 & 0.21 & 8\%  & \textbf{4.35} & 1.56 & 0.94 & 0.47 \\
C3LM         & 42\% & 1.69 & 1.14 & 0.78 & 0.40 & 24\% & \underline{4.21} & \underline{2.06} & \underline{1.29} & \underline{0.65} \\ 
\midrule
\multicolumn{11}{c}{\textit{Liquid Foundation Models}} \\
\midrule
LFM2-2.6B                      & 0\% & 0.00 & 0.00 & 0.00 & 0.00 & 0\% & 0.00 & 0.00 & 0.00 & 0.00 \\ 
LFM2-2.6B-MMAI Gym, MT, \texttt{/no\_think}     & \textbf{94\%} & 1.15 & 0.64 & 0.41 & 0.21 & \textbf{90\%} & 2.85 & 1.56 & 1.00 & 0.51 \\
LFM2-2.6B-MMAI Gym, MT, \texttt{/think}     & \textbf{94\%} & 1.45 & 0.84 & 0.55 & 0.28 & \underline{89\%} & 3.10 & 1.79 & 1.18 & 0.60 \\
LFM2-2.6B-MMAI Gym, ST, \texttt{/no\_think}  & 92\% & 1.15 & 0.64 & 0.41 & 0.21 & 89\% & 2.91 & 1.61 & 1.05 & 0.53 \\
LFM2-2.6B-MMAI Gym, ST, \texttt{/think}   & \underline{93\%} & 1.43 & 0.90 & 0.60 & 0.30 & 88\% & 3.22 & 1.86 & 1.24 & 0.631 \\
\midrule
LFM2-24B-A2B                      & 0\% & 0.00 & 0.00 & 0.00 & 0.00 & 0\% & 0.00 & 0.00 & 0.00 & 0.00 \\ 
LFM2-24B-A2B-MMAI Gym, ST, \texttt{/think}     & 70\% & \underline{1.73} & \textbf{1.38} & \textbf{1.09} & \textbf{0.63} & 83\% & 3.32 & \textbf{2.27} & \textbf{1.76} & \textbf{1.02} \\ 
\bottomrule
\end{tabular}
}
\caption{\textbf{Single-Step Retrosynthesis benchmarking results.} Columns report the following metrics. \textbf{Unique}: fraction of unique valid reactant sets among samples. \textbf{Max}: per-target maximum ChemCensor score averaged over targets. \textbf{Av.\ PT-Top-K CC}: per-target average ChemCensor score over top-K unique predictions. We highlight best values with \textbf{bold} and second best with \underline{underline}.}
\label{tab:ssrs_res_ext}
\end{table*}

\begin{table*}[t]
\centering
\small
\resizebox{1\textwidth}{!}{
    \begin{tabular}{ll|c|c|cc|cc|cc|c|c}
\hline
\multirow{3}{*}{\textbf{Group}} & \multirow{3}{*}{\textbf{Task}} & \multirow{3}{*}{\textbf{Metric}} &
\multicolumn{5}{c|}{\textbf{LFM-2.6B}} &
\multicolumn{2}{c|}{\textbf{LFM2-24B-A2B}} &
\multirow{3}{*}{\textbf{TxGemma-27B}} &
\multirow{3}{*}{\textbf{TDC SOTA}} \\
\cline{4-10}
& & &
\multirow{2}{*}{\textbf{Base}} &
\multicolumn{2}{c|}{\textbf{MT}} &
\multicolumn{2}{c|}{\textbf{ST}} &
\multirow{2}{*}{\textbf{Base}} &
\multirow{2}{*}{\textbf{ST}} & & \\
\cline{5-8}
& & &
& \texttt{/no\_think} & \texttt{/think} & \texttt{/no\_think} & \texttt{/think} & & & & \\
\hline
\multirow{18}{*}{ADME/PK}
& Caco2 Wang              & MAE ($\downarrow$)     & 0.550 & 0.365 & 0.396 & 0.356 & \underline{0.347} & 0.703 & 0.36 & 0.401 & \textbf{0.256} \\
& Lipophilicity AZ        & MAE ($\downarrow$)     & 0.953 & 0.454 & \underline{0.441} & 0.457 & \textbf{0.434} & 1.001 & 0.57 & 0.538 & 0.456 \\
& Solubility AqSolDB      & MAE ($\downarrow$)     & 2.066 & 0.833 & \underline{0.812} & 0.836 & \underline{0.812} & 2.602 & 0.84 & 0.907 & \textbf{0.741} \\
& PPBR AZ                 & MAE ($\downarrow$)     & 10.130 & \underline{7.722} & 7.755 & 7.917 & 8.006 & 11.029 & 8.62 & 9.048 & \textbf{7.440} \\
& VDss Lombardo           & Spearman ($\uparrow$)  & 0.131 & 0.585 & 0.596 & 0.589 & 0.591 & 0.043 & \underline{0.63} & 0.559 & \textbf{0.713} \\
& Half Life Obach         & Spearman ($\uparrow$)  & -0.023 & 0.418 & 0.380 & 0.402 & 0.381 & -0.02 & \underline{0.46} & 0.458 & \textbf{0.576} \\
& Clearance Microsome AZ  & Spearman ($\uparrow$)  & -0.020 & 0.542 & 0.554 & 0.566 & 0.531 & -0.072 & \underline{0.59} & 0.462 & \textbf{0.630} \\
& Clearance Hepatocyte AZ & Spearman ($\uparrow$)  & -0.036 & 0.372 & 0.377 & 0.370 & 0.356 & -0.124 & \underline{0.48} & 0.260 & \textbf{0.536} \\
& Pgp Broccatelli         & AUROC ($\uparrow$)     & 0.585 & 0.887 & 0.852 & 0.896 & 0.863 & — & 0.91 & \underline{0.937} & \textbf{0.938} \\
& HIA Hou                 & AUROC ($\uparrow$)     & 0.428 & 0.981 & 0.987 & 0.977 & 0.987 & — & \underline{0.99} & 0.988 & \textbf{0.993} \\
& BBB Martins             & AUROC ($\uparrow$)     & 0.366 & \underline{0.930} & 0.922 & \underline{0.930} & 0.924 & — & \textbf{0.96} & 0.908 & 0.924 \\
& Bioavailability Ma      & AUROC ($\uparrow$)     & 0.516 & \underline{0.752} & \underline{0.752} & 0.746 & 0.751 & — & 0.74 & 0.694 & \textbf{0.942} \\
& CYP2C9 Substrate CM     & AUPRC ($\uparrow$)     & 0.346 & 0.427 & 0.427 & 0.426 & \underline{0.444} & — & 0.4 & 0.438 & \textbf{0.474} \\
& CYP2C9 Veith            & AUPRC ($\uparrow$)     & 0.372 & 0.774 & 0.752 & 0.772 & 0.751 & — & \underline{0.78} & 0.683 & \textbf{0.859} \\
& CYP2D6 Substrate CM     & AUPRC ($\uparrow$)     & 0.366 & 0.649 & 0.631 & 0.658 & 0.630 & — & 0.69 & \underline{0.711} & \textbf{0.736} \\
& CYP2D6 Veith            & AUPRC ($\uparrow$)     & 0.206 & 0.631 & 0.582 & 0.636 & 0.585 & — & 0.61 & \underline{0.683} & \textbf{0.790} \\
& CYP3A4 Substrate CM     & AUROC ($\uparrow$)     & 0.571 & \textbf{0.719} & 0.686 & \underline{0.711} & 0.681 & — & 0.7 & 0.691 & 0.667 \\
& CYP3A4 Veith            & AUPRC ($\uparrow$)     & 0.477 & \underline{0.865} & 0.845 & 0.862 & 0.855 & — & 0.86 & 0.854 & \textbf{0.916} \\
\hline
\multirow{4}{*}{Tox}
& LD50 Zhu                & MAE ($\downarrow$)     & 0.799 & 0.706 & 0.709 & 0.711 & 0.703 & 0.823 & 0.67 & \underline{0.627} & \textbf{0.552} \\
& DILI                    & AUROC ($\uparrow$)     & 0.612 & \underline{0.950} & 0.941 & 0.947 & 0.937 & — & 0.89 & 0.886 & \textbf{0.956} \\
& hERG                    & AUROC ($\uparrow$)     & 0.679 & 0.779 & 0.789 & 0.784 & 0.787 & — & 0.83 & \textbf{0.885} & \underline{0.880} \\
& AMES                    & AUROC ($\uparrow$)     & 0.522 & 0.805 & 0.792 & 0.800 & 0.782 & — & 0.81 & \underline{0.816} & \textbf{0.871} \\
\hline
\end{tabular}
}
\caption{\textbf{ADMET Properties Prediction results on TDC.} We report the same base and LFM2-2.6B-MMAI model results after MMAI Gym. ST is single-category RFT, including all TDC ADMET tasks. TDC SOTA is the specialist models leaderboard. Best is \textbf{bold}, second best is \underline{underlined}.}
\label{tab:tdc_res_ext}
\end{table*}

\subsection{MuMO-Instruct Results}
\label{app:mumo_res}

\textbf{Task, Setup, and Metrics.} We evaluate our model's capability in multi-objective lead optimization using the MuMO-Instruct benchmark \citep{liu2025fgbench}. Unlike standard single-property tasks, MuMO-Instruct requires the model to modify a starting molecule to simultaneously improve multiple conflicting properties while retaining structural similarity. We report results on five core molecular property combinations: BDP (BBBP, DRD2, PlogP), BDQ (BBBP, DRD2, QED), BPQ (BBBP, PlogP, QED), DPQ (DRD2, PlogP, QED), and the four-property task BDPQ. Performance is measured using three metrics: Success Rate (SR), the percentage of generated molecules that satisfy all property optimization constraints; Similarity (Sim), the Tanimoto similarity between the input and generated optimized molecule; and Relative Improvement (RI), quantifying the magnitude of property enhancement. For each test example, we generate and aggregate $20$ prediction samples using independently sampled prompt templates and molecular-input augmentations.

\textbf{Baselines.} We benchmark LFM2-2.6B-MMAI against a comprehensive suite of models: Foundational Chemistry LLMs; Task-specific non-LLMs and LLMs; Generalist LLMs, which represent the current state-of-the-art for this benchmark.

\textbf{Summary of Results.} Table~\ref{tab:mumo_res_ext} shows that MMAI Gym training substantially improves both LFM2 models on multi-objective lead optimization. The MMAI-trained LFM2-2.6B variants achieve the strongest Success Rate across most three-property tasks, consistently outperforming proprietary LLMs, foundational chemistry models, and generalist approaches. On the most challenging four-property BDPQ setting, LFM2-24B-A2B-MMAI attains the best Success Rate overall, demonstrating strong optimization capabilities even under the most demanding multi-objective constraints.

Across all metrics, the MMAI-trained models demonstrate consistent strength. While some specialized baselines achieve higher structural similarity or relative improvement, these often come at the cost of substantially lower Success Rates. This indicates a tendency to either preserve the input molecule without sufficient optimization or make modifications that fail to satisfy all property constraints simultaneously. The MMAI-trained LFM2 models strike an effective balance: they achieve leading Success Rates while maintaining competitive similarity and improvement, demonstrating their ability to perform effective multi-property optimization while retaining reasonable structural similarity to the starting molecules.

Overall, these comprehensive results demonstrate that MMAI Gym training substantially strengthens multi-objective molecular optimization capabilities and enables LFM2 models to compete with, and often surpass, both specialized task-specific models trained exclusively on this task and much larger proprietary LLMs.

\subsection{FGBench Benchmark Results}
\label{app:fgbench_res}
\textbf{Task, Setup, and Metrics}. The FGBench~\cite{liu2025fgbench} is an instruction-following benchmark for functional-group editing, a key task that reflects molecular property reasoning. Each example in the dataset consists of a reference molecule and a textual description of its modification (e.g., adding or removing named functional groups at specified positions) and asks the model to predict how a specified property changes under that edit. The dataset supports three types of questions: the effect of modifying a single functional group, the effect of modifying multiple groups, and the question of predicting a specified property for the target molecule given a similar reference molecule and its value of the property. Each group of questions comes in True / False and numeric value prediction variants. We calculate accuracy (\textbf{Acc}) for the binary tasks and \textbf{RMSE} for regression. For each test example, we generate and aggregate $15$ prediction samples using independently sampled prompt templates and molecular-input augmentations.

\textbf{Baselines.} We compare the performance of two LFM2-2.6B-MMAI models against several baseline models: a) the original LFM-2 model prior to MMAI gym; b) proprietary general-purpose LLMs; c) open-weight general-purpose LLMs; and d) chemical specialist LLMs.

\textbf{Summary of results.} As shown in Table~\ref{tab:fgbench_res_ext}, MMAI Gym training leads to substantial and consistent improvements over the base LFM2 models across all three FGBench task types and for both boolean classification and numeric value prediction. The LFM2-MMAI variants achieve best or second-best performance across the majority of metrics, demonstrating strong functional-group reasoning capabilities.

The gains are particularly pronounced on the more challenging Interaction and Comparison settings, where MMAI-trained models substantially outperform base models in both accuracy and RMSE. This indicates that MMAI Gym training strengthens the model's ability to reason about complex functional-group effects and their influence on molecular properties. Among the variants, \texttt{/think} mode consistently achieves the best RMSE scores for value prediction tasks, while both \texttt{/think} and \texttt{/no\_think} variants demonstrate strong boolean classification performance.

Notably, all LFM2-MMAI models maintain perfect output validity while achieving superior accuracy and lower prediction errors compared to baselines. The models substantially outperform chemistry-specialized LLMs and remain competitive with or superior to much larger proprietary models. The combination of strong performance, perfect validity, and efficient model size demonstrates that MMAI Gym training effectively enhances functional-group reasoning and molecular property prediction capabilities.

\textbf{Error analysis.} For a focused comparison of model scales, we analyze the three boolean FGBench tasks. Across $5,696$ examples, the 24B model succeeds on $524$ cases ($9.2$\%) that the 2.6B variant answers incorrectly. These tasks require comparing molecular properties before and after an edit, or reasoning about a pair of closely related molecules. In most of these cases, the smaller model incorrectly estimates the relevant properties and consequently reaches the wrong conclusion. Less frequently, it applies the molecular edit incorrectly and then reasons consistently about the resulting, but incorrect, structure. This suggests that the primary limitation of the smaller model lies in property estimation rather than instruction following or edit execution.

\subsection{Single-Step Retrosynthesis Results}
\label{app:ssrs_res}

\textbf{Task, Setup, and Metrics.} We evaluate our models for the ability to perform a single-step retrosynthesis (SSRS) task \citep{choe2025retrosyntheticcrosstalk}, a standard task related to full-scale synthesis planning; on the conventional and curated USPTO-50K test benchmark \citep{liu2017uspto50ktest} based on public data; and the URSA-expert-2026 benchmark \citep{zagribelnyy2026chemcensor}, which contains out-of-distribution (OOD) target molecules with undisclosed answers. ChemCensor (CC) \cite{zagribelnyy2026chemcensor}, a recently introduced data-driven metric for evaluating chemical plausibility based on synthetic precedents, as well as aggregated metrics based on CC, were used to benchmark the models' performance in the SSRS task. For each target molecule, we generate $15$ candidate reactants sets and evaluate the resulting valid unique reactant sets.

\textbf{Baselines.} We compare the performance of the LFM2-2.6B-MMAI models against several baseline tiers: a) proprietary general-purpose (GP) LLMs; b) open-weight GP LLMs; c) chemical generalist LLMs, and d) the original LFM2-2.6 models prior to the domain-specific training.

\textbf{Summary of results.} As shown in Table~\ref{tab:ssrs_res_ext}, MMAI Gym training dramatically improves retrosynthesis capabilities, transforming the base LFM2 models from zero performance to achieving leading results across both benchmarks. The MMAI-trained models generate substantially higher fractions of unique valid predictions compared to all baseline categories, including proprietary LLMs and chemical specialists.

On the aggregated ChemCensor metrics, LFM2-24B-A2B-MMAI achieves the best performance on both benchmarks across all top-K settings, surpassing all proprietary LLMs, open-weight models, and chemical specialists. The LFM2-2.6B-MMAI variants also demonstrate strong plausibility scores while maintaining very high diversity. Across variants, reasoning-enabled (\texttt{/think}) inference consistently improves ChemCensor-based plausibility metrics compared to non-reasoning modes, indicating that explicit chemical reasoning helps the model select more credible and synthetically feasible disconnections.

While some specialized models achieve high maximum ChemCensor scores on individual targets, they generate far fewer unique predictions, suggesting overfitting to common patterns. The MMAI-trained LFM2 models strike an effective balance, achieving both high diversity and strong average plausibility. Overall, these results demonstrate that MMAI Gym training substantially strengthens retrosynthetic reasoning, enabling LFM2 models to outperform both general-purpose and chemistry-specialized baselines on this challenging synthesis planning task.

\subsection{TDC Benchmark Results}
\label{app:tdc_res}

\textbf{Task, Setup, and Metrics.} We evaluate models on the Therapeutics Data Commons (TDC) benchmark \citep{huang2021tdc}, a standardized platform for drug discovery tasks. We specifically focus on the absorption, distribution, metabolism, excretion (ADME), pharmacokinetics (PK), and toxicity groups, which require predicting pharmacological properties such as absorption (Caco2 cells monolayer permeability, human intestine absorption (HIA)), distribution (Blood-brain barrier (BBB) penetration, plasma protein binding in rat plasma), metabolism (cytochrome P450 enzymes), and safety (hERG inhibition, Ames genetoxicity test). Following the standard TDC protocol, we utilize task-specific metrics including Mean Absolute Error (MAE) for regression tasks, Spearman’s Correlation for pharmacokinetics, and AUROC or AUPRC for classification tasks. The baselines are the established TDC State-of-the-Art (SOTA) Leaderboard ~\citep{huang2021tdc} and a much larger general-purpose scientific model trained on TDC, TxGemma-27B~\citep{wang2025txgemma}. For each test example, we generate and aggregate $10$ prediction samples using independently sampled prompt templates and molecular-input augmentations.

\textbf{Baselines.} We compare the performance of LFM2-2.6B-MMAI models against several baseline tiers: a) LFM2-2.6B: the original autoregressive hybrid model without domain-specific training to isolate the impact of our MMAI Gym; b) TxGemma-27B \citep{wang2025txgemma}: a heavyweight 27-billion parameter language model tailored for therapeutics; c) TDC SOTA: the highest recorded performance on the official TDC leaderboard for each respective task.

\textbf{Summary of Results.} As shown in Table \ref{tab:tdc_res_ext}, LFM2-2.6B-MMAI consistently improves over the base LFM2-2.6B across essentially all ADME/PK and toxicity tasks, highlighting the benefit of MMAI Gym training. Across the Liquid model variants, the best-performing configurations are competitive with substantially larger models: they surpass TxGemma-27B on multiple ADME/PK task and are comparable on others. While we achieve performance on par with or superior to heavyweight LLMs, our model does not yet consistently beat the specialized TDC SOTA in regression tasks like Clearance or VDss, where specialist non-LLM models still maintain a lead. Across settings, multi-task RFT (MT) training is generally more robust and competitive across the full benchmark suite, while single-task RFT (ST) training can yield the strongest results on particular datasets. We further observe that enabling thinking often provides a small but consistent lift on harder endpoints, whereas no reasoning model is typically comparable on easier tasks and offers a lower-compute alternative with minimal loss in performance.

\textbf{Error analysis.} We further compare the 2.6B and 24B models on the $13$ TDC classification tasks, where each example asks whether a molecule exhibits a particular property, such as blood-brain barrier penetration. Of the $10,498$ evaluated examples, $1,195$ ($11.4$\%) are answered correctly only by the 24B model. The Ames and three Veith CYP tasks account for $90.3$\% of these cases, whereas HIA contributes none, suggesting that additional model capacity is most beneficial for the more challenging classification endpoints. Qualitative inspection further indicates that the 2.6B model is more likely to fail on larger molecules, where relevant structures and their interactions may be harder to resolve.

\subsection{Spatial Molecular Distribution Learning Results}
\label{app:distribution_learning_3d}

\begin{table*}[t]
\centering
\small
\resizebox{1\textwidth}{!}{
    \begin{tabular}{c|c|c|c|c|c|c|c}
\toprule
\multirow{2}{*}{\textbf{Group}} & \multirow{2}{*}{\textbf{Metrics}} & \multicolumn{4}{c|}{\textbf{Baselines}} & \multicolumn{2}{c}{\textbf{LFM-MMAI-Gym}} \\
\cline{3-8}
  & & \textbf{MolDiff} & \textbf{OpenLLaMA} & \textbf{nach0} & \textbf{nach0-pc} & \textbf{2.6B} & \textbf{24B-A2B} \\
\midrule
\multirow{4}{*}{Basic}
& Validity and Connectivity ($\uparrow$) & \textbf{99.3\%} & 95.0\% & \underline{98.6\%} & 97.8\% & 97.3\% & 96.3\% \\
& Uniqueness ($\uparrow$) & 92.8\% & \textbf{99.8\%} & \textbf{99.8\%} & 99.2\% & 99.0\% & \underline{99.5\%} \\
& Novelty ($\uparrow$) & \underline{97.7\%} & 85.0\% & \textbf{98.0\%} & 96.2\% & 96.1\% & 96.5\% \\
& Diversity ($\uparrow$) & \underline{0.763} & 0.745 & 0.739 & \textbf{0.781} & 0.738 & 0.716 \\
\midrule
\multirow{3}{*}{Druglikeness}
& QED ($\uparrow$) & 0.679 & 0.673 & 0.667 & \textbf{0.771} & \underline{0.698} & 0.673 \\
& SA ($\uparrow$) & \textbf{0.875} & 0.808 & 0.851 & \underline{0.872} & 0.845 & 0.846 \\
& Lipinski ($\uparrow$) & \underline{4.981} & 4.972 & 4.938 & \textbf{4.992} & 4.971 & 4.963 \\
\midrule
\multirow{3}{*}{3D substr.}
& JS. bond lengths ($\downarrow$) & 0.436 & 0.257 & 0.148 & 0.193 & \textbf{0.116} & \underline{0.122} \\
& JS. bond angles ($\downarrow$) & 0.181 & 0.136 & \underline{0.096} & 0.100 & \textbf{0.094} & 0.098 \\
& JS. dihedral angles ($\downarrow$) & 0.198 & \underline{0.110} & \underline{0.110} & 0.132 & \textbf{0.101} & 0.139 \\
\midrule
\multirow{4}{*}{Bonds}
& JS. \# bonds per atoms ($\downarrow$)  & 0.121 & \textbf{0.064} & \underline{0.111} & 0.233 & 0.118 & 0.131 \\
& JS. freq. bond types ($\downarrow$) & 0.170 & \textbf{0.031} & 0.046 & 0.052 & \underline{0.040} & 0.066 \\
& JS. freq. bond pairs ($\downarrow$) & 0.153 & \textbf{0.028} & \underline{0.037} & 0.040 & 0.046 & 0.080 \\
& JS. freq. bond triplets ($\downarrow$) & 0.137 & \textbf{0.034} & \underline{0.046} & 0.054 & 0.052 & 0.085 \\
\midrule
\multirow{3}{*}{Rings}
& JS. \# rings ($\downarrow$) & \underline{0.079} & \textbf{0.033} & 0.56 & 0.270 & 0.081 & 0.111 \\
& JS. \# n-sized rings ($\downarrow$) & 0.102 & \underline{0.024} & 0.026 & 0.058 & \textbf{0.019} & 0.025 \\
& \# Intersecting rings ($\uparrow$) & \underline{8} & \underline{8} & \underline{8} & \textbf{9} & \textbf{9} & \underline{8} \\
\bottomrule
\end{tabular}
}
\caption{\textbf{Spatial molecular distribution learning results.} We report generation quality, drug-likeness, and distributional metrics over 3D substructures, bonds, and rings. Best is \textbf{bold}, second best is \underline{underlined}.}
\label{tab:distribution_learning_3d_ext}
\end{table*}

\begin{figure*}
  \centering
  \includegraphics[width=.98\linewidth]{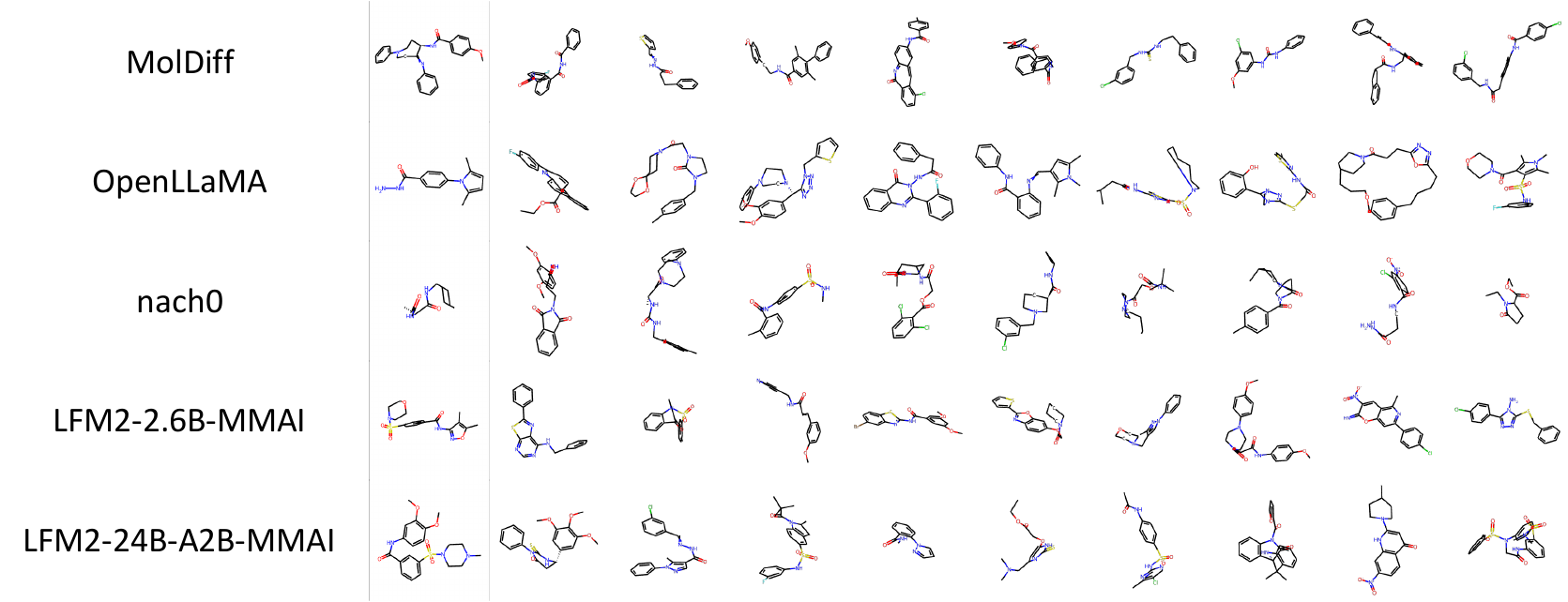}
  \caption{Samples for \textbf{spatial molecular distribution learning} task}
  \label{fig:distr_learn_samples}
\end{figure*}

\textbf{Task, Setup, and Metrics.} We evaluate the model's ability to generate structurally and physically plausible 3D molecular objects through the \textit{spatial molecular distribution learning} task. The goal is to generate novel 3D molecular structures whose distribution is close to the ground-truth data distribution. Following \cite{peng2023moldiff}, we use the high-quality GEOM-Drugs dataset \cite{axelrod2022geom}, which provides gold-standard conformer ensembles generated with metadynamics in CREST \cite{pracht2020automated}. We assess generated molecules from several perspectives, including basic generation metrics such as validity and novelty, as well as drug-likeness and distributional divergences over 3D substructures, bonds, and rings. For each evaluated model, we generate $10{,}000$ molecular structures for computing the reported generation-quality and distributional metrics.

\textbf{Baselines.} We compare our model against prior baselines for 3D molecular generation reported in \cite{livne2024nacho}, including language models and diffusion-based approaches designed to model the distribution of spatial molecular structures.

\textbf{Summary of Results.} We present results in Table~\ref{tab:distribution_learning_3d_ext}, with qualitative examples shown in Figure~\ref{fig:distr_learn_samples}. We omit the base LFM2 models without MMAI Gym training from this comparison, since in our experiments they generally fail to follow the required output format, making reliable evaluation on this benchmark infeasible.

The MMAI Gym-trained LFM variants demonstrate strong performance across multiple evaluation dimensions, achieving best or second-best results on $8$ out of $17$ metrics. Most notably, the models excel at capturing 3D geometric distributions: LFM2-2.6B-MMAI achieves the best performance on all three 3D substructure metrics (bond lengths, bond angles, and dihedral angles), substantially outperforming all baselines including specialized diffusion models. This indicates that MMAI Gym training enables accurate modeling of local 3D molecular geometry. Both MMAI-trained models maintain high generation quality with strong uniqueness and novelty, demonstrating their ability to generate diverse and novel molecular structures. 

Overall, the results demonstrate that MMAI Gym training successfully enables LFM2 models to generate physically plausible 3D molecular structures with particularly strong geometric fidelity, while maintaining competitive performance on drug-likeness and generation quality metrics.

\subsection{Pocket-Conditioned Generation Results}
\label{app:pocket_gen}

\begin{table*}[t]
\centering
\small
\resizebox{1\textwidth}{!}{
    \begin{tabular}{l|cc|cc|cc|cc|cc}
\toprule
\multirow{3}{*}{Model}& \multicolumn{6}{c|}{\textbf{Molecular Validity}} & \multicolumn{4}{c}{\textbf{Pocket}} \\ 
\cline{2-11}
&
\multicolumn{2}{c|}{Parsed} &
\multicolumn{2}{c|}{PB Intra} &
\multicolumn{2}{c|}{Int. Energy} &
\multicolumn{2}{c|}{PB Inter} &
\multicolumn{2}{c}{UniDock} \\
& \textbf{CD} & \textbf{PL} & \textbf{CD} & \textbf{PL} & \textbf{CD} & \textbf{PL} & \textbf{CD} & \textbf{PL} & \textbf{CD} & \textbf{PL} \\
\midrule
\multicolumn{11}{c}{\textit{Diffusion Models}} \\
\midrule
PocketXMol   & \textbf{100\%} & \textbf{100\%} & \textbf{95\%} & \textbf{97\%} & \underline{99\%}  & \textbf{99\%} & \textbf{98\%} & \textbf{99\%} & \underline{-8.1} & -7.0 \\
SeFMol       & \textbf{100\%} & \textbf{100\%} & 51\% & 55\% & 95\%  & 95\% & 0\%  & 1\%  & -6.2 & -5.4 \\
DiffPharma   & \textbf{100\%} & \textbf{100\%} & 80\% & 66\% & \textbf{100\%} & 92\% & 90\% & \underline{91\%} & -4.9 & -5.1 \\
MolSnapper   & \textbf{100\%} & \textbf{100\%} & \underline{89\%} & 85\% & 83\%  & 80\% & \underline{92\%} & \underline{91\%} & \textbf{-8.6} & \textbf{-8.2} \\
IPDiff       & -     & \textbf{100\%} & -    & 54\% & -     & 94\% & -    & 1\%  & -    & -5.5 \\
BindDM       & \textbf{100\%} & \textbf{100\%} & 43\% & 44\% & 90\%  & 92\% & 0\%  & 1\%  & -5.3 & -4.4 \\
DiffSBDD     & \underline{99\%}  & \underline{99\%}  & 70\% & 73\% & 92\%  & 92\% & 90\% & \underline{91\%} & -5.8 & -4.9 \\
PMDM         & 88\%  & 87\%  & 72\% & 70\% & 81\%  & 78\% & 1\%  & 1\%  & -7.7 & \underline{-7.3} \\
\midrule
\multicolumn{11}{c}{\textit{Proprietary LLMs}} \\
\midrule
GPT 5.5        & \textbf{100\%} & \textbf{100\%} & \textbf{95\%} & \underline{95\%} & 96\%  & 96\% & 36\% & 42\% & -4.7 & -4.4 \\
GPT 5.4        & 73\%  & 74\%  & 9\%  & 11\% & 44\%  & 47\% & 1\%  & 1\%  & -5.7 & -5.3 \\
Opus 4.7       & 85\%  & 80\%  & 84\% & 78\% & 84\%  & 80\% & 17\% & 20\% & -4.5 & -4.2 \\
Opus 4.6       & 77\%  & 74\%  & 66\% & 64\% & 73\%  & 73\% & 0\%  & 0\%  & -6.3 & -5.7 \\
Sonnet 4.6     & 86\%  & 92\%  & 64\% & 69\% & 75\%  & 84\% & 1\%  & 1\%  & -5.4 & -4.9 \\
Gemini 3.1-Pro & \underline{99\%}  & \textbf{100\%} & 86\% & 87\% & 96\%  & \underline{97\%} & 32\% & 42\% & -5.2 & -4.7 \\
Grok 4.1       & 98\%  & \underline{99\%}  & 84\% & 83\% & 87\%  & 87\% & 3\%  & 2\%  & -4.9 & -4.6 \\
\midrule
\multicolumn{11}{c}{\textit{Open-weight LLMs}} \\
\midrule
Qwen 3.5 397B & 97\%  & 96\%  & 57\% & 59\% & 67\%  & 71\% & 6\%  & 6\%  & -5.0 & -4.5 \\
DeepSeek V3.2 & 66\%  & 69\%  & 9\%  & 11\% & 32\%  & 33\% & 1\%  & 1\%  & -5.3 & -4.8 \\
GLM 5         & 54\%  & 41\%  & 42\% & 29\% & 44\%  & 31\% & 8\%  & 8\%  & -4.9 & -4.4 \\
\midrule
\multicolumn{11}{c}{\textit{Liquid Foundation Models}} \\
\midrule
LFM2-2.6B-MMAI & 93\%  & 95\%  & 56\% & 52\% & 79\%  & 83\% & 0\%  & 0\%  & -7.3 & -7.0 \\
LFM2-24B-A2B-MMAI & 98\%  & 98\%  & 55\% & 57\% & 85\%  & 81\% & 0\%  & 0\%  & -7.4 & -7.0 \\
\bottomrule
\end{tabular}
}
\caption{\textbf{Pocket-Conditioned generation results.} Best is \textbf{bold}, second best is \underline{underlined}.}
\label{tab:pocket_conditioned_ext}
\end{table*}

\begin{figure*}
  \centering
  \includegraphics[width=.98\linewidth]{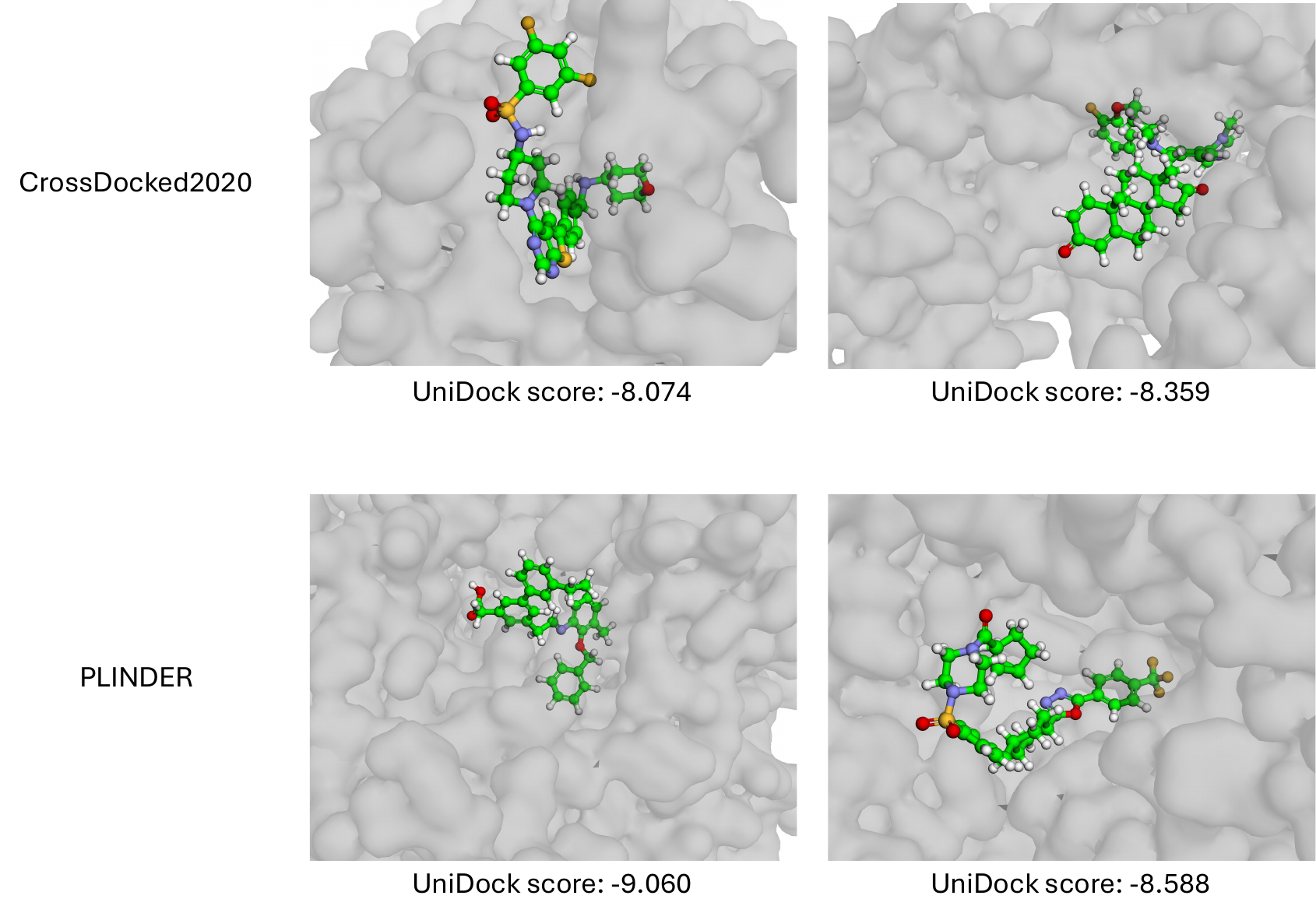}
  \caption{Samples for \textbf{pocket-conditioned ligand generation} task from LFM2-2.6B-MMAI model.}
  \label{fig:pocked_conditioned_samples}
\end{figure*}

\textbf{Task, Setup, and Metrics.} We evaluate the MMAI-Gym-trained LFM2 models on pocket-conditioned 3D ligand generation using 3D-Fit benchmark~\cite{macdougall2026threedfit}, where the model is asked to generate a ligand structure compatible with a given binding pocket. We assess generated molecules along two complementary axes: \textbf{3D molecular validity} and \textbf{3D condition success}. For validity, we report whether the generated structure can be successfully parsed and sanitized, along with PoseBusters-based intramolecular plausibility checks and internal energy filters. For pocket-conditioning success, we report docking-based pocket compatibility using UniDock score and intermolecular PoseBusters checks that evaluate whether the generated ligand is plausibly placed in the protein pocket. Lower docking score is better, while higher success rates are better for parsing, validity, and PoseBusters metrics. For each evaluated model, we generate one ligand per target pocket.

\textbf{Baselines.} We compare the MMAI-Gym-trained LFM2 models against a set of strong baselines covering both general-purpose and specialist approaches for 3D molecular generation. These include general-purpose LLMs, as well as specialist 3D diffusion models designed for ligand generation in protein pockets. This comparison allows us to evaluate how well the adapted LFM2 models transfer language-based molecular reasoning to a structure-based drug design setting.

\textbf{Summary of Results.} We present results in Table~\ref{tab:pocket_conditioned_ext}, with qualitative examples shown in Figure~\ref{fig:pocked_conditioned_samples}. The MMAI Gym-trained LFM2 models demonstrate strong performance on this challenging task, achieving the best UniDock docking scores among all language models. Most notably, both LFM2-MMAI variants substantially outperform all proprietary and open-weight LLM baselines on UniDock scores-the primary metric for measuring predicted binding affinity-and approach the performance of specialist diffusion models. An interesting comparison emerges with GPT-5.5, which represents a contrasting strategy: GPT-5.5 produces highly valid and geometrically plausible molecules with near-perfect parsing rates and best intramolecular validity checks, but achieves only moderate UniDock binding affinity scores. In contrast, the MMAI Gym-trained models optimize more directly for binding affinity, generating high-scoring ligands while maintaining competitively lower validity rates. This suggests MMAI Gym training successfully adapts LFM2 to prioritize the key objective of structure-based drug design-strong predicted binding affinity-rather than playing it safe with conservative, highly valid but less optimized structures. While specialist diffusion models still lead on overall binding scores, the LFM2-MMAI models surpass several diffusion baselines-a notable step for language-model-based structure-based drug design.

\section{Ablation Study}
\label{app:ablation}

We study four main design choices where comparable configurations are available:

(i) the effect of MMAI Gym training, measured against the corresponding base LFM2 models; 

(ii) multi- vs single-task (MT vs.\
ST); 

(iii) reasoning (\texttt{/think} vs.\ \texttt{/no\_think}); 

(iv) model scale (2.6B vs 24B);

Since not all configurations are evaluated on every benchmark, we restrict each comparison to available settings.

\paragraph{Effect of MMAI Gym training.} MMAI Gym training consistently improves performance across tasks. On MuMO-Instruct (Table~\ref{tab:mumo_res_ext}), adapted models satisfy substantially more optimization constraints than their base counterparts while retaining competitive similarity. On FGBench (Table~\ref{tab:fgbench_res_ext}), training improves functional-group classification and property prediction, especially for interaction and comparison questions, while preserving output validity. The largest gains occur for specialized outputs: base models fail to produce useful retrosynthesis predictions or reliable 3D representations, whereas adapted models generate plausible reactants and perform competitively on spatial tasks (Tables~\ref{tab:ssrs_res_ext}, \ref{tab:distribution_learning_3d_ext}, and~\ref{tab:pocket_conditioned_ext}). Improvements on TDC (Table~\ref{tab:tdc_res_ext}) further demonstrate benefits for discriminative property prediction.

\paragraph{Multi-task versus single-task RFT.} Multi-task (MT) and single-task (ST) RFT achieve similar overall performance with complementary strengths. Multi-task is generally more robust across tasks and performs well on multi-property optimization, while single-task sometimes improves highly constrained optimization and specialized FGBench results. In retrosynthesis, multi-task tends to preserve prediction diversity, whereas single-task can improve chemical plausibility. Neither strategy consistently dominates on TDC: multi-task provides broader robustness, while single-task occasionally performs better on individual endpoints. Overall, multi-task is preferable for transfer across tasks, whereas single-task can offer modest gains through specialization.

\paragraph{Effect of inference-time reasoning.} The value of \texttt{/think} is task-dependent. It consistently improves FGBench, especially numerical questions involving interactions between functional groups, and generally yields more chemically plausible retrosynthetic disconnections while preserving prediction diversity. In contrast, its effect on TDC is mixed, likely because many property-prediction tasks resemble direct mappings from molecular structure to scalar or categorical outputs. Thus, \texttt{/no\_think} is a lower-compute option for straightforward property prediction, whereas \texttt{/think} is most useful for compositional reasoning, molecular editing, and synthesis planning.

\paragraph{Effect of model scale.} Scaling produces task-specific rather than uniform gains. LFM2-2.6B-MMAI remains highly competitive on molecular optimization and local 3D geometry, while LFM2-24B-A2B-MMAI performs better on some highly constrained tasks, including four-property optimization and retrosynthetic plausibility. The two sizes achieve similar pocket-conditioned docking performance, with the larger model providing somewhat better parsing but no consistent advantage in validity. Overall, additional capacity is most beneficial for multi-constraint or plausibility-sensitive tasks, while the smaller model offers a stronger performance-efficiency trade-off.

\paragraph{Overall conclusion.} MMAI Gym training is the primary driver of improvement. Multi-task and single-task training are complementary: multi-task favors broad transfer and robustness, whereas single-task can improve specialized tasks. Inference-time reasoning is most valuable for tasks requiring chemical reasoning or search and is less reliable for direct property prediction. Larger LFM2-24B-BMMAI-Gym model help on some difficult multi-constraint tasks, but do not consistently outperform smaller LFM2-2.6B-MMAI-Gym.

\clearpage
\onecolumn
\begin{longtable}{
    @{}
    >{\raggedright\arraybackslash}p{0.22\textwidth}
    >{\raggedright\arraybackslash}p{0.34\textwidth}
    >{\raggedleft\arraybackslash}p{0.14\textwidth}
    >{\raggedleft\arraybackslash}p{0.18\textwidth}
    @{}
}
\caption{\textbf{Dataset and corpus statistics.}
The table reports the number of processed training objects and the mean number of tokens per object where these statistics are currently available. }
\label{tab:dataset_statistics}\\

\toprule
\textbf{Task family} &
\textbf{Dataset} &
\textbf{Objects} &
\textbf{Mean tokens/object} \\
\midrule
\endfirsthead

\multicolumn{4}{c}{
\tablename~\thetable{}: Dataset and corpus statistics (continued)
}\\
\toprule
\textbf{Task family} &
\textbf{Dataset} &
\textbf{Objects} &
\textbf{Mean tokens/object} \\
\midrule
\endhead

\midrule
\multicolumn{4}{r}{Continued on the next page}\\
\endfoot

\bottomrule
\endlastfoot

\multicolumn{4}{c}{\textit{Molecular graph and sequence tasks}} \\
\midrule
Mol. optimization
    & MuMO-Instruct~\cite{dey2025mumoinstruct}
    & $745{,}260$
    & $3{,}486$ \\
Functional groups
    & FGBench~\cite{liu2025fgbench}
    & $618{,}790$
    & $1{,}588$ \\
Retrosynthesis
    & CREED~\cite{zagribelnyy2026chemcensor}
    & $2{,}838{,}069$
    & $1{,}171$ \\
Molecular generation
    & MOSES~\cite{polykovskiy2020moses}
    & $1{,}584{,}663$
    & $706$ \\
Molecular description
    & MolTextNet~\cite{zhu2025moltextnet}
    & $2{,}448{,}055$
    & $3{,}246$ \\

\midrule
\multicolumn{4}{c}{
    \textit{TDC~\cite{huang2021tdc}: ADME and pharmacokinetic property prediction}
} \\
\midrule
Property prediction
    & Caco2 Wang
    & $637$
    & $3{,}468$ \\
Property prediction
    & Lipophilicity AZ
    & $2{,}940$
    & $3{,}247$ \\
Property prediction
    & Solubility AqSolDB
    & $6{,}987$
    & $3{,}104$ \\
Property prediction
    & PPBR AZ
    & $1{,}952$
    & $3{,}515$ \\
Property prediction
    & VDss Lombardo
    & $791$
    & $2{,}995$ \\
Property prediction
    & Half Life Obach
    & $466$
    & $3{,}460$ \\
Property prediction
    & Clearance Microsome AZ
    & $771$
    & $3{,}551$ \\
Property prediction
    & Clearance Hepatocyte AZ
    & $849$
    & $3{,}736$ \\
Property prediction
    & Pgp Broccatelli
    & $852$
    & $3{,}481$ \\
Property prediction
    & HIA Hou
    & $404$
    & $3{,}309$ \\
Property prediction
    & BBB Martins
    & $1{,}421$
    & $3{,}467$ \\
Property prediction
    & Bioavailability Ma
    & $448$
    & $3{,}447$ \\
Property prediction
    & CYP2C9 Substrate CM
    & $468$
    & $3{,}508$ \\
Property prediction
    & CYP2C9 Veith
    & $8{,}464$
    & $3{,}412$ \\
Property prediction
    & CYP2D6 Substrate CM
    & $466$
    & $3{,}211$ \\
Property prediction
    & CYP2D6 Veith
    & $9{,}191$
    & $3{,}460$ \\
Property prediction
    & CYP3A4 Substrate CM
    & $468$
    & $3{,}267$ \\
Property prediction
    & CYP3A4 Veith
    & $8{,}629$
    & $3{,}167$ \\

\midrule
\multicolumn{4}{c}{\textit{TDC~\cite{huang2021tdc}: toxicity prediction}} \\
\midrule
Property prediction
    & LD50 Zhu
    & $5{,}169$
    & $3{,}202$ \\
Property prediction
    & DILI
    & $332$
    & $3{,}384$ \\
Property prediction
    & hERG
    & $458$
    & $3{,}543$ \\
Property prediction
    & AMES
    & $5{,}094$
    & $3{,}130$ \\

\midrule
\multicolumn{4}{c}{\textit{Spatial molecular tasks}} \\
\midrule
3D mol. generation
    & GEOM-Drugs~\cite{axelrod2022geom}
    & $243{,}473$
    & $1{,}233$ \\
3D mol. generation
    & ZINC~\cite{irwin2005zinc}
    & $164{,}962$
    & $1{,}846$ \\
Property prediction
    & QMUGS~\cite{isert2022qmugs}
    & $1{,}793{,}694$
    & $4{,}307$ \\
Pocket-conditioned generation
    & CrossDocked2020~\cite{francoeur2020crossdocked}
    & $100{,}000$
    & $8{,}182$ \\ 
\midrule
\multicolumn{4}{c}{\textit{Protein tasks}} \\
\midrule
Protein folding
    & AlphaFoldDB~\cite{varadi2023alphafolddb}
    & $542{,}378$
    & $9{,}188$ \\
Protein folding
    & SCOPE~\cite{chandonia2021scope}
    & $3{,}914$
    & $2{,}406$ \\
Peptides modeling
    & BC40~\cite{berman2000pdb}
    & $30{,}858$
    & $8{,}641$ \\
Cyclic peptides modeling
    & CREMP~\cite{grambow2024cremp}
    & $35{,}198$
    & $1{,}993$ \\

\midrule
\multicolumn{4}{c}{
    \textit{Cross-domain and instruction-following tasks}
} \\
\midrule
Language-molecules 
    & L+M-24~\cite{edwards2024lm24}
    & $160{,}560$
    & $1{,}291$ \\
Protein instructions
    & ProteinLMBench~\cite{shen2024proteinlmbench}
    & $893{,}996$
    & $3{,}669$ \\

\midrule

\end{longtable}
\clearpage
\twocolumn

\end{document}